\documentclass[12pt]{article}

\usepackage[T1]{fontenc}

\usepackage[compress]{cite}
\usepackage{IEEEtrantools,stackengine}
\stackMath
\usepackage{amsmath}

\usepackage{authblk}
\usepackage{fullpage}
\usepackage{amssymb,amsmath}
\usepackage{adjustbox}
\usepackage{graphicx}
\usepackage{longtable}

\usepackage{tcolorbox}
\tcbuselibrary{xparse,skins,breakable}

\usepackage{inconsolata}

\usepackage{listings}
\lstdefinelanguage{json}{
    breaklines=true,
}
\lstdefinestyle{modernjson}{
    language=json,
    basicstyle=\linespread{1}\ttfamily\small, % modern + single spaced
    keywordstyle=\color{blue!70!black},
    stringstyle=\color{green!50!black},
    commentstyle=\color{red!60!black},
    showstringspaces=false,
    numbers=left,
    numberstyle=\tiny\color{gray},
    stepnumber=1,
    tabsize=2,
    breaklines=true
}

\tcbset{%
    basestyle/.style={enhanced,breakable,
        fonttitle=\bfseries,
        arc=1mm,
        boxrule=1.8pt,
        coltitle=white,
        left=5pt,
        right=15pt
        },
    defstyle/.style={basestyle,
                    colback=gray!10,
                    colframe=gray!150,},
    theostyle/.style={basestyle,colback=blue!10!white},
}

\DeclareTColorBox[auto counter]{exlisting}{ o O{} t\label g }{%
    defstyle,
    IfValueTF={#1}{title={Extended Listing~\thetcbcounter\ -- #1}}{title=Extended Listing~\thetcbcounter},
    IfBooleanTF={#3}{label=#4}{},#2}
    
\usepackage{siunitx}
\usepackage{csquotes}

\usepackage{xspace}

\usepackage{booktabs}
\usepackage{longtable}
\usepackage{multirow}

\usepackage[dvipsnames, table]{xcolor}

\usepackage[left]{lineno}

\definecolor{darkgreen}{rgb}{0.0, 0.5, 0.0}

\definecolor{lightyellow}{HTML}{FFE699}
\definecolor{red_revision}{HTML}{FF0000}

\usepackage{setspace}
\usepackage[unicode=true]{hyperref}
\hypersetup{breaklinks=true,
            pdfborder={0 0 0},
            allcolors=black}
\usepackage[%
  sort&compress
]{cleveref}
  \Crefname{appendix}{Supplement}{Supplements}
  \Crefname{figure}{Fig.}{Fig.}

\usepackage{times}
\usepackage{enumerate}
\usepackage{enumitem}
\usepackage{float}

\usepackage{subfig}
\usepackage{graphicx}
\usepackage{caption}

\usepackage{rotating}
\usepackage{tabularx}
\usepackage{dcolumn}
\usepackage{pdflscape}
\usepackage{rotating}

\usepackage{array}

\usepackage{titlesec}

\titleformat{\subsubsection}
  {\normalfont\itshape}{\thesubsubsection}{1em}{}

\titleformat{\subsection}
  {\normalfont\bfseries}{\thesubsection}{1em}{}

\makeatletter
\renewcommand{\fps@figure}{H}         % default {tbp}
\renewcommand{\fps@table}{H}         % default {tbp}
\makeatother

\allowdisplaybreaks
\usepackage{eurosym}
\usepackage{comment}

\usepackage{ntheorem}
\theoremseparator{:}

\usepackage{makecell}

\usepackage[margin=1in]{geometry}
\usepackage{tcolorbox}
\usepackage{xcolor}

\begin{document}

%%%%%%%%%%%%%%%%%%%%%%%%%%%%%%%%%%%%%%%%%%%%%%%%%%%%%%%%%%%%%%%%%%%%%%%%%%%%%%
% Title page
% Information-seeking large language models enhance diagnostic reasoning in medicine 
% MedClarify: Atumoated diagnostic reasoning through information-seeking  large language models in medicine

\title{\centering\LARGE\singlespacing 
MedClarify: An information-seeking AI agent for medical diagnosis with case-specific follow-up questions
}

\renewcommand\Affilfont{\fontsize{9}{10.8}\selectfont}

% % Author names
\author[1]{Hui Min Wong}
\author[1,2,3]{Philip Heesen}
\author[1,2]{Pascal Janetzky}
\author[3]{Martin Bendszus}
\author[1,2]{Stefan Feuerriegel}

\affil[1]{LMU Munich, Munich, Germany}
\affil[2]{Munich Center for Machine Learning, Germany}
\affil[3]{Department of Neuroradiology, Heidelberg University, Heidelberg, Germany}

\date{}

\maketitle

%%%%%%%%%%%%%%%%%%%%%%%%%%%%%%%%%%%%%%%%%%%%%%%%%%%%%%%%%%%%%%%%%%%%%%%%%%%%%%%%%%%%

% \newpage
%\linenumbers

\begin{abstract}\normalfont
\noindent
Large language models (LLMs) are increasingly used for diagnostic tasks in medicine. In clinical practice, the correct diagnosis can rarely be immediately inferred from the initial patient presentation alone. Rather, reaching a diagnosis often involves systematic history taking, during which clinicians reason over multiple potential conditions through iterative questioning to resolve uncertainty. This process requires considering differential diagnoses and actively excluding emergencies that demand immediate intervention. Yet, the ability of medical LLMs to generate informative follow-up questions and thus reason over differential diagnoses remains underexplored. Here, we introduce MedClarify, an AI agent for information-seeking that can generate follow-up questions for iterative reasoning to support diagnostic decision-making. Specifically, MedClarify computes a list of candidate diagnoses analogous to a differential diagnosis, and then proactively generates follow-up questions aimed at reducing diagnostic uncertainty. By selecting the question with the highest expected information gain, MedClarify enables targeted, uncertainty-aware reasoning to improve diagnostic performance. In our experiments, we first demonstrate the limitations of current LLMs in medical reasoning, which often yield multiple, similarly likely diagnoses, especially when patient cases are incomplete or relevant information for diagnosis is missing. We then show that our information-theoretic reasoning approach can generate effective follow-up questioning and thereby reduces diagnostic errors by $\sim$27 percentage points (p.p.) compared to a standard single-shot LLM baseline. Altogether, MedClarify offers a path to improve medical LLMs through agentic information-seeking and to thus promote effective dialogues with medical LLMs that reflect the iterative and uncertain nature of real-world clinical reasoning.
\end{abstract}

% \flushbottom
% \maketitle
% \thispagestyle{empty}

%\begin{center}
%\begin{tabular}{p{14.5cm}}
%\small
%\noindent\textbf{Keywords}: keyword, ...
%\end{tabular}
%\end{center}

\sloppy
\raggedbottom

%%%%%%%%%%%%%%%%%%%%%%%%%%%%%%%%%%%%%%%%%%%%%%%%%%%%%%%%%%%%%%%%%%%%%%%%%%%%%%

\newpage
\section*{Introduction}
\label{sec:introduction}

% LLMs in medicine

Large language models (LLMs) are increasingly integrated into clinical workflows \cite{thirunavukarasu_large_2023, williams_evaluating_2024,clusmann_future_2023}, especially to assist with diagnostic decision-making \cite{moor_foundation_2023, tu_towards_2025, chen_enhancing_2025, tang_medagents_2024, kanjee_accuracy_2023,eriksen_use_2024, goh_large_2024, hager_evaluation_2024, goh_gpt-4_2025, liu_generalist_2025, olmo_assessing_2024, chen_rarebench_2024, abdullahi_learning_2024,johri_evaluation_2025, truhn_artificial_2026}. By synthesizing patient information and suggesting probable diagnoses, LLMs may not only support clinicians in managing complex and rare diseases, but also prevalent conditions where fast and accurate diagnosis is critical \cite{liu_generalist_2025, olmo_assessing_2024, chen_rarebench_2024, abdullahi_learning_2024}. This potential is particularly important given that diagnostic errors remain a major source of harm in medicine, leading to an estimated 800,000 cases of death or permanent disability each year in the U.S. alone \cite{newman-toker_burden_2024}.

% context: importance of information-seeking in clincial practice

A central challenge in medical diagnosis is that the information obtained during initial history taking is often incomplete, creating substantial uncertainty for diagnostic decision-making \cite{alegria_how_2008, graber_diagnostic_2005}. This extends beyond the patient history alone: physical examination findings, laboratory results, and imaging may be unavailable, pending or only available through active clinical inquiry. Patients typically report only their primary presenting symptom, but oftentimes leave out important details such as symptom timeline, contextual factors, relevant medical history, or medication intake which clinicians must actively inquire \cite{bhise_defining_2018, wagner_chief_2006, raven_comparison_2013, engebretsen_uncertainty_2016, han_varieties_2011}. Missing or ambiguous information broadens the differential diagnosis and increases the risk of misdiagnosis, especially when some conditions cannot yet be ruled out \cite{cook_higher_2020}. To resolve this uncertainty, clinicians commonly ask targeted follow-up questioning to clarify symptoms and distinguish among competing diagnoses  \cite{leeds_teaching_2020, gatens-robinson_clinical_1986, elstein_medical_1978, mandin_helping_1997}. For example, when evaluating unexplained chest pain, a clinician may probe specifically for exertional symptoms, radiation patterns, or risk factors to differentiate cardiac from non-cardiac cause \cite{gulati_2021_2021}. Similarly, when assessing a patient suspected of type 2 diabetes, physicians will also inquire about dietary habits and physical activity \cite{kalyani_diagnosis_2025}. By continually seeking additional information, physicians can systematically narrow down the set of possible diseases and update the probability of various differential diagnoses \cite{gatens-robinson_clinical_1986, leeds_teaching_2020, mandin_helping_1997,elstein_medical_1978}. Automating this iterative, question-driven reasoning process represents a promising direction for AI-assisted diagnosis.

% research gap: LLMs don't do information-seeking

However, current medical LLMs are not designed to follow the information-seeking approach that is common in clinical practice. Instead, existing medical LLM systems \cite{wu_medical_2024,seki_assessing_2025,zhuang_learning_2025,takita_systematic_2025,jin_what_2021,gao_leveraging_2025,shieh_assessing_2024,kanjee_accuracy_2023,liu_generalist_2025,olmo_assessing_2024,chen_rarebench_2024,zhou_uncertainty-aware_2025} are typically designed to follow a static input-output paradigm with a single-shot prediction: given the description of a patient case, the LLM produces a single diagnosis or a list of potential diagnosis. Yet, many of the generated diagnoses are incorrect \cite{li_mediq_2024, wang_healthq_2025}, and the correct diagnosis often cannot be identified without additional follow-up questions. Hence, a common limitation \cite{hager_evaluation_2024} of medical LLMs is that these cannot identify relevant but missing information and then generate case-specific follow-up questions to reduce diagnostic uncertainty and guide reasoning toward the correct diagnosis. While there are some methods to generate confirmatory questions for a single diagnosis \cite{mahajan_cognitive_2025, elston_confirmation_2020}, similar approaches that identify the correct diagnosis out of a pool of multiple candidate diagnoses through follow-up questions are missing.  

% what we do

To overcome these limitations of current LLM approaches, we here introduce MedClarify, an AI agent for information-seeking that generates follow-up questions during a diagnostic workflow (see Figure~\ref{fig:flowchart}). MedClarify first computes a set of candidate diagnoses, analogous to a differential diagnosis, and then proactively proposes follow-up questions aimed at reducing diagnostic uncertainty. For this, we propose a novel information gain formulation---called diagnostic expected information gain (DEIG)---to quantify how much a follow-up question reduces the entropy of the potential, case-specific diagnosis distribution, which we then integrate into MedClarify in a Bayesian framework. As a result, MedClarify iteratively identifies which question would most effectively reduce diagnostic uncertainty to arrive at the correct diagnosis. Thereby, MedClarify offers a path to improve medical LLMs through active information-seeking and to thus promote effective dialogues consistent with the iterative and uncertain nature of real-world clinical reasoning.

We evaluate MedClarify through an agentic framework using several medical datasets for diagnostic benchmarking, namely, the NEJM Image Challenge \cite{noauthor_image_nodate}, MediQ \cite{li_mediq_2024}, and MedQA \cite{jin_what_2021}, and assess its information-seeking capabilities. We implement MedClarify using an open-source LLM, namely, Llama 3.3-70B \cite{touvron_llama_2023} as the primary backbone, but later also provide a comparison with alternative LLM backbones. Our experiments show that off-the-shelf LLMs struggle when key information in patient cases is missing; even state-of-the-art LLMs generate frequent misdiagnoses, which increases diagnostic errors by up to 10--15 p.p. This also reveals a weakness of current benchmarking practices, which typically assume complete and fully specified patient cases---an assumption that rarely holds in routine clinical settings. In clinical practice, important information is often missing or unknown at first presentation, making single-shot predictions prone to diagnostic errors. In contrast, we find that MedClarify substantially improves diagnostic accuracy by generating informative follow-up questions that reduce diagnostic uncertainty and guide reasoning toward the correct diagnosis. Across our experiments, we find improvements in diagnostic accuracy by up to $\sim$27 p.p. compared to single-shot LLM prediction on incomplete patient cases, which correspond to reductions in diagnostic errors by $\sim$48\%.

%%%%%%%%%%%%%%%%%%%%%%%%%%%%%%%%%%%%%%%%%%%%%%%%%%%%%%%%%%%%%%%%%%%%%%%%%%%%%%
\newpage
\section*{Results}
\label{sec:results}

\subsection*{MedClarify for information-seeking in medical LLMs}

\begin{figure}
\centering
\includegraphics[width=1\linewidth]{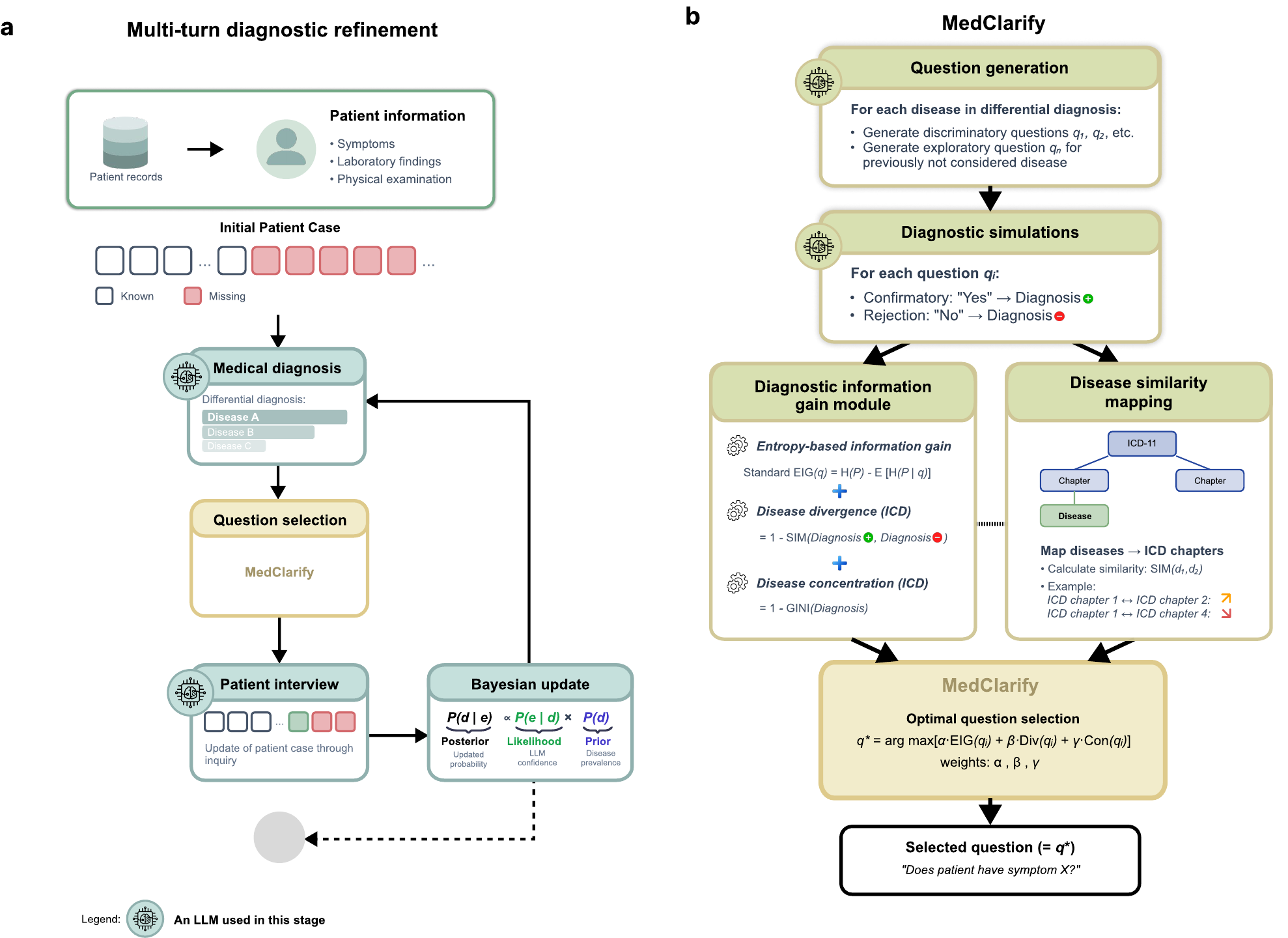}
\caption{\textbf{Overview of MedClarify, an AI agent for medical diagnosis that optimizes the question selection strategy.} \textbf{a},~The system generates a set of differential diagnoses with corresponding confidence scores, which are then adjusted using a Bayesian update that combines knowledge from old and new evidence. Based on this improved differential diagnosis, MedClarify selects the optimal follow-up question through optimizing information gain, thus reducing uncertainty. This question is then used to interview the patient, and the patient's response is integrated into the patient's case. This iterative diagnostic refinement process is repeated until specific thresholds are met, and eventually outputs a final diagnosis. \textbf{b},~The question selection strategy and technical overview of MedClarify.}
\label{fig:flowchart}
\end{figure}

\begin{figure}
\begin{center}
\includegraphics[width=.85\linewidth]{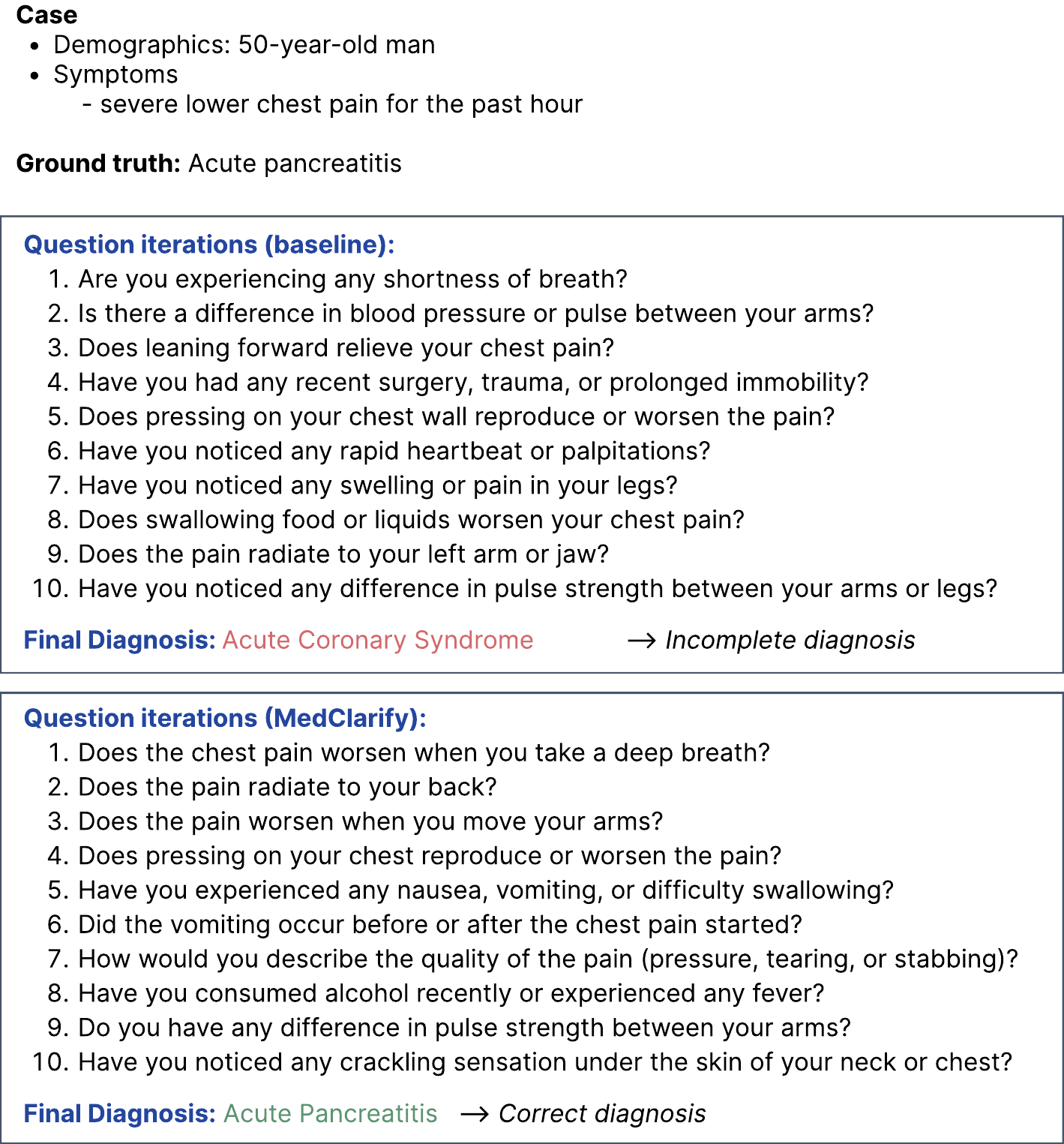}
\end{center}

\caption{\textbf{Example with proactive, case-specific follow-up questions for medical diagnosis.} The figure presents an example patient case with clinical information (top), as well an the diagnostic dialogue generated by the na{\"i}ve multi-turn baseline system (middle) and by MedClarify (bottom) for the same patient case. Additional dialogues are in Extended Figure~\ref{exfig:ex-effective-ex1} and ~\ref{exfig:ex-effective-ex2}. }
\label{fig:ex-effective}
\end{figure}

MedClarify is an agentic, information-seeking system for LLMs that takes a patient case as input and then identifies the most informative follow-up question to refine a differential diagnosis (Figure~\ref{fig:flowchart}a). For this, MedClarify iteratively asks a series of case-specific follow-up questions during patient interview in order to identify the correct diagnosis from a set of candidate diagnoses. At its core, MedClarify implements a principled, data-driven query-selection strategy that prioritizes questions expected to reduce diagnostic uncertainty and rule out competing diagnoses (Figure~\ref{fig:flowchart}b). To quantify diagnostic uncertainty, MedClarify estimates the entropy \cite{he_entropy_2024} of the LLM’s distribution over candidate diagnoses, but where we extend the standard entropy formulation to incorporate the semantic relatedness of diagnoses such that related conditions contribute less to overall uncertainty than entirely distinct ones. Specifically, the MedClarify system maps the predicted medical diagnoses onto ICD codes (i.e., the International Classification of Diseases [ICD-11], 11th Revision \cite{drosler_icd-11_2021}) and then measures the semantic proximity between potential conditions. This accounts for how diseases are medically related and thus enables MedClarify to generate follow-up questions that can more effectively rule out entire branches of related conditions. An example dialogue with case-specific follow-up questions is shown Figure~\ref{fig:ex-effective} (further examples are in Extended Figure~\ref{exfig:ex-effective-ex1} and \ref{exfig:ex-effective-ex2}).

% eval framework

To evaluate the performance of MedClarify, we simulate patient interviews using an agentic evaluation framework. The evaluation framework allows MedClarify to perform iterative diagnostic reasoning by asking follow-up questions, updating its differential diagnosis, and converging toward the correct condition through interaction with a simulated patient and their clinical records. The evaluation framework comprises components: (i)~a \textsc{Patient} agent that responds to questions; (ii)~a \textsc{Doctor} agent that performs diagnostic reasoning using MedClarify and queries the \textsc{Patient} agent to retrieve more information; (iii)~a \textsc{Update} agent that updates the patient case with the newly retrieved information; and (iv)~an \textsc{Evaluator} agent that assesses the correctness of the final diagnosis while accounting for variations in medical terminology. Together, these components create a controlled setting for evaluating how well MedClarify gathers missing information and improves diagnostic accuracy.

\begin{figure}
\centering
\rlap{\raisebox{0.97\height}{\hspace{0.05\linewidth}\textbf{\textsf{a}}}}%
\includegraphics[width=0.97\linewidth]{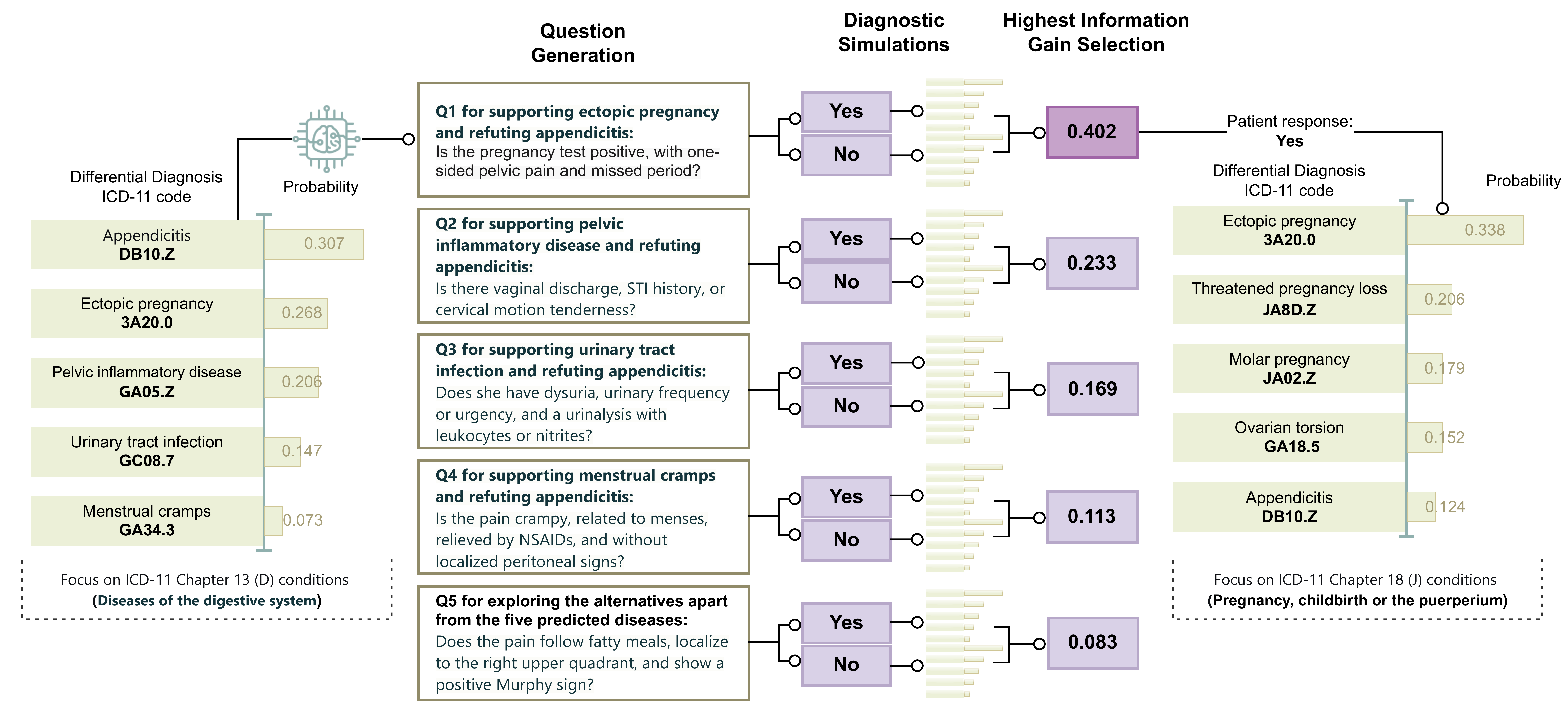}

\rlap{\raisebox{0.97\height}{\hspace{0.05\linewidth}\textbf{\textsf{b}}}}%
\includegraphics[width=0.97\linewidth]{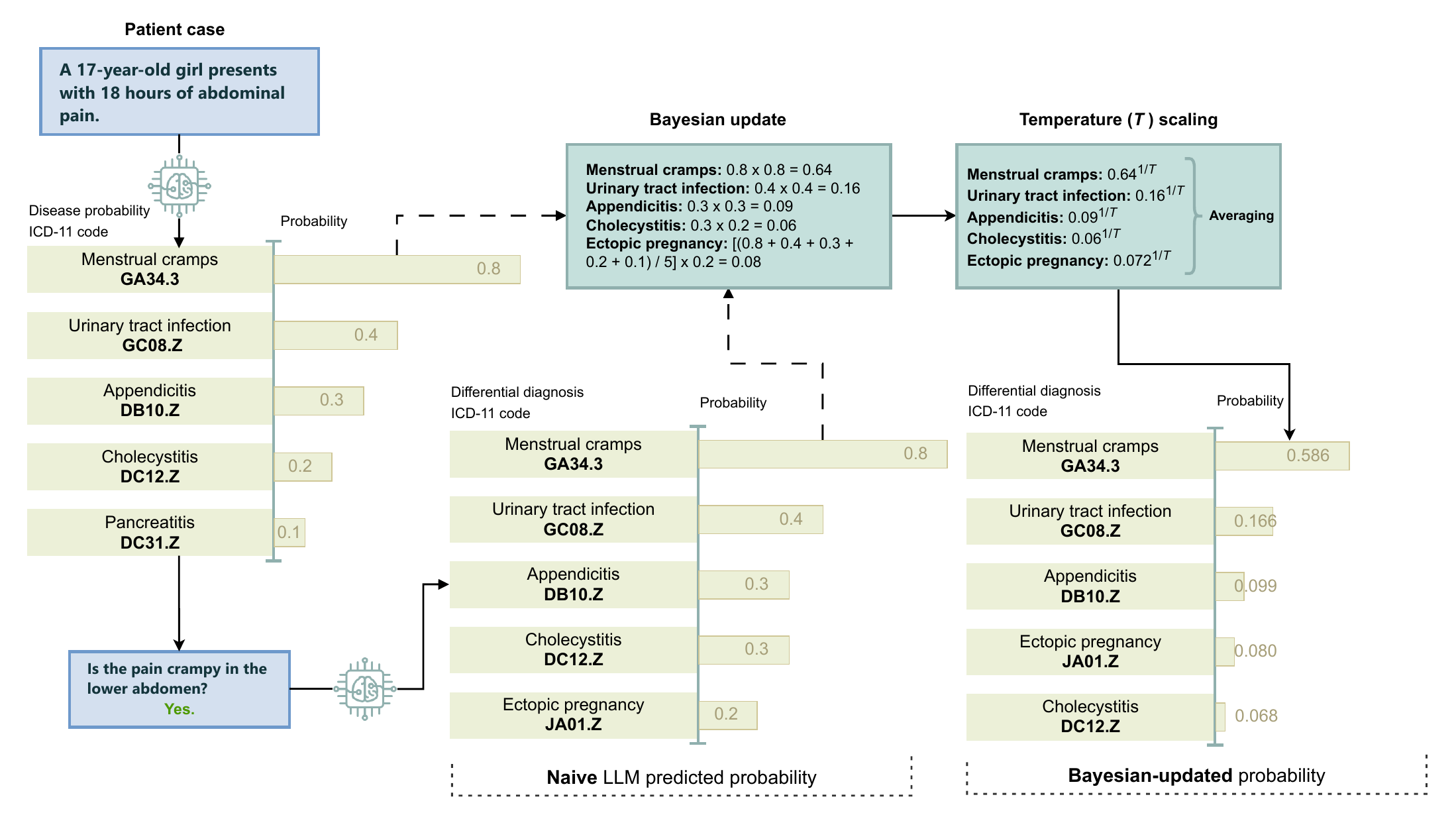}
\vspace{0.8em}

\caption{\textbf{Overview of Bayesian update and question selection in MedClarify.} \textbf{a},~Question selection strategy in MedClarify can shift LLM focus to alternative condition groups, encouraging to rule out a wider range of diseases for a more reliable final diagnosis. The process begins by generating targeted questions for each predicted disease to refute the highest-probability condition and confirm low-ranked alternatives, plus an additional explorative question. Afterwards, the system selects the question with the highest information gain. In this example, although the initial hypothesis was related to ICD-11 Chapter 13 (disease of the digestive system), MedClarify selects a question that effectively confirms Chapter 18 (pregnancy, childbirth, or the puerperium), leading to the correct diagnostic pathway, which the baseline system would have missed. \textbf{b},~Bayesian update prevents static entropy by incorporating historical diagnoses and temperature scaling.}
\label{fig:bayesian}
\end{figure}

MedClarify uses an entropy–based information-seeking strategy (which we call diagnostic expected information gain [DEIG]) to identify follow-up questions that most effectively reduce diagnostic uncertainty. For each candidate question, DEIG evaluates how the distribution over possible diagnoses is expected to change and selects the question that rules out incorrect alternatives while shifting probability toward the true diagnosis. Intuitively, DEIG favors questions that meaningfully narrow the differential diagnosis—questions whose answers eliminate broad sets of conditions rather than only distinguishing between minor variants (Figure~\ref{fig:bayesian}a). For example, when the LLM is uncertain between pneumonia, pulmonary embolism, and myocardial ischemia, a question about pleuritic chest pain can collectively rule out several non-pulmonary causes, thereby concentrating probability on the correct diagnostic branch. Because MedClarify incorporates ICD-based semantic similarity, it treats closely related diagnoses, such as different pneumonia subtypes, as belonging to the same broader group. This prevents inflated uncertainty from minor wording differences and allows follow-up questions to eliminate entire clusters of related conditions. At each step, MedClarify applies a Bayesian update to the diagnostic distribution so that the newly acquired evidence systematically reweights the probabilities rather than leaving them static (Figure~\ref{fig:bayesian}b).

MedClarify is evaluated on a total of 469 patient cases from the NEJM Image Challenge($n = 170$ cases) \cite{noauthor_image_nodate}, MediQ ($n = 129$) \cite{li_mediq_2024}, and MedQA ($n = 170$) \cite{jin_what_2021}. These cases span a broad range of diseases across eight medical specialties: cardiology, pulmonology, gastroenterology, neurology, dermatology, endocrinology, hematology/oncology, and urology/nephrology (Extended Figure~\ref{fig:dist}). We assess MedClarify using a state-of-the-art open-source LLM backbone, namely, Llama 3.3-70B \cite{touvron_llama_2023}. The model is evaluated with MedClarify and compared against a standard single-shot approach, where the LLM directly generates a diagnosis from the initial case description without asking any follow-up questions.

\subsection*{Diagnostic performance degrades under incomplete patient cases}

To assess how missing clinical information (such as missing symptoms, missing laboratory tests, etc.) affects LLM diagnostic accuracy, we conducted feature masking experiments using the patient cases from the NEJM, MediQ, and MedQA datasets. We categorized clinical information from the patient cases into six feature categories: symptoms, social history, past medical history, physical examination, laboratory tests, and imaging results. Then, we masked each feature category individually to simulate scenarios where incomplete patient information is presented to the diagnostic LLM. Diagnostic accuracy on each masked case was then compared against performance on the corresponding complete case.

\begin{figure}
%\centering
\begin{minipage}[t]{0.98\linewidth}
    \centering

\begin{tabular}{ll}
    \llap{\textbf{\textsf{a}}\hspace{0.5em}} 
    &
    \llap{\textbf{\textsf{b}}\hspace{0.5em}}%
    \\
    \includegraphics[width=.45\linewidth]{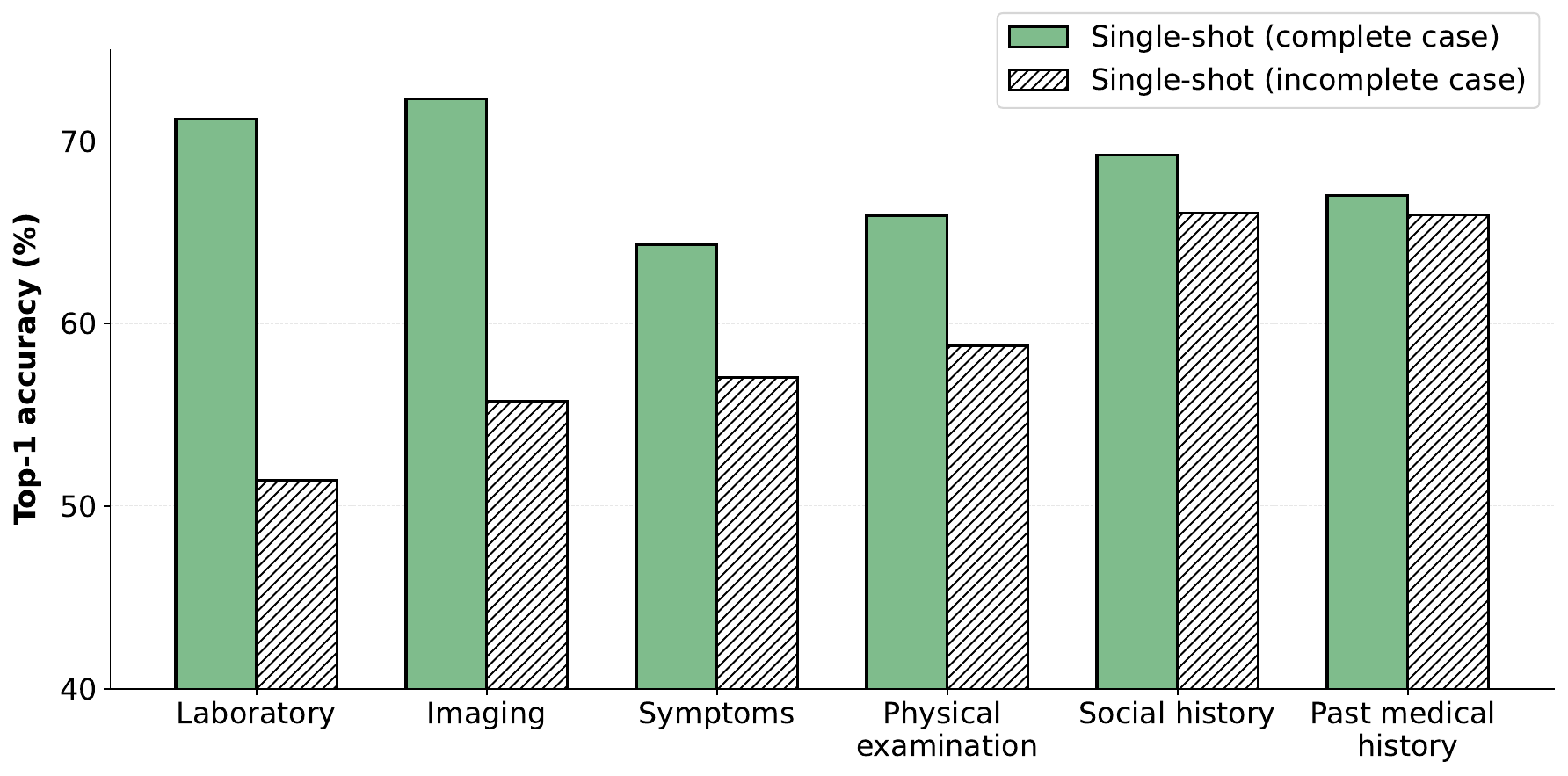}
    \vspace{1em} 
    &
    \includegraphics[width=.45\linewidth]{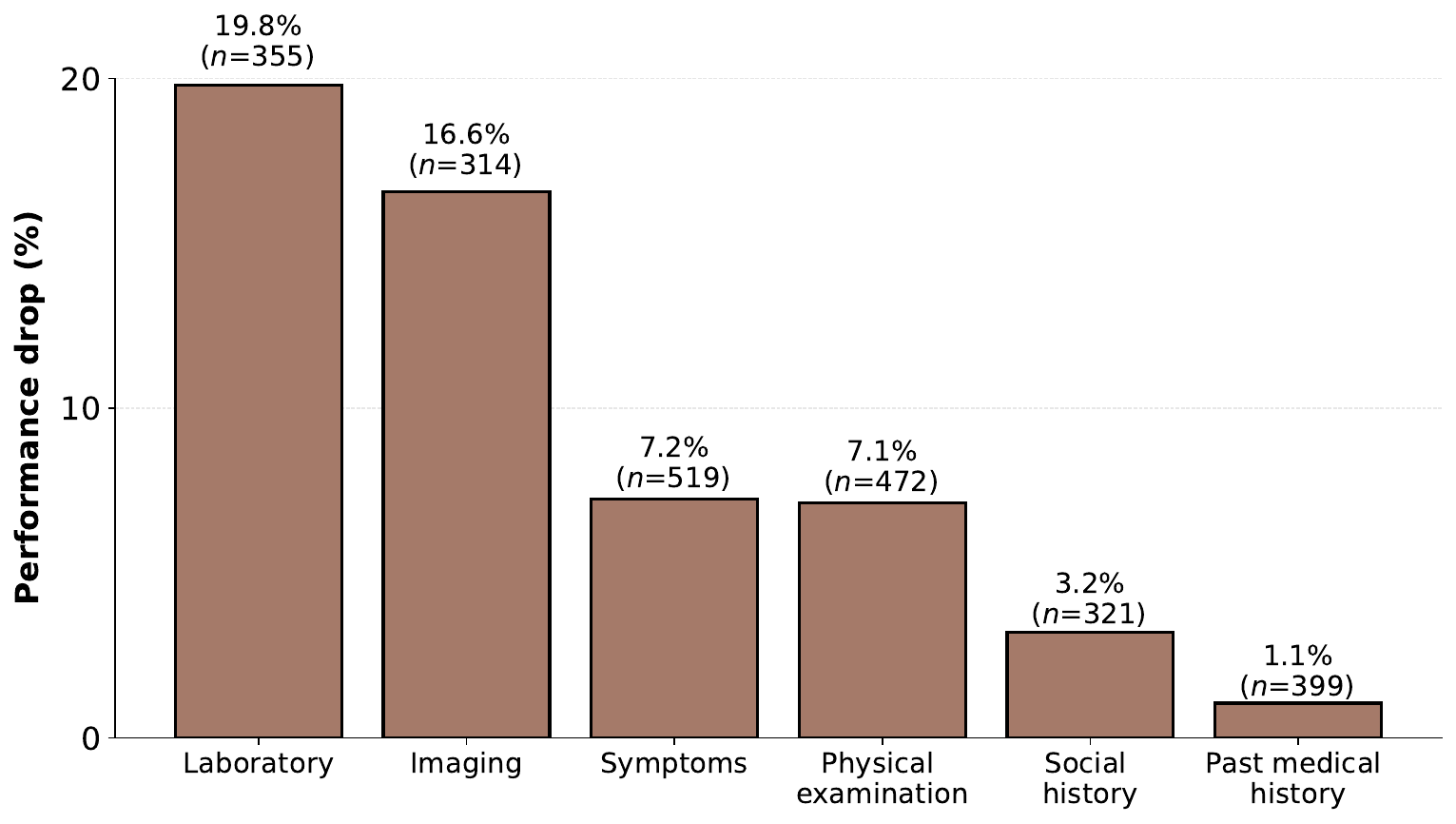}
\end{tabular}
\end{minipage}
\caption{\textbf{Decrease in LLM diagnostic accuracy under incomplete patient cases.} \textbf{a},~Diagnostic accuracy of Llama-3.3-70B in single-shot without inquiries on complete cases (green) and on cases with a single clinical feature masked (striped), evaluated across all datasets (NEJM, MediQ, and MedQA) for each feature category (laboratory findings, imaging results, symptoms, physical examinations, social history, and past medical history). \textbf{b},~Decrease in diagnostic accuracy (in p.p.) for masked cases relative to the corresponding complete cases across different feature categories (This shows the subtraction of striped bars from solid green bars across feature categories in \textbf{a}). 
}
\label{fig:accuracy-drop}
\end{figure}

Masking clinical information led to systematic degradation in diagnostic accuracy (Figure~\ref{fig:accuracy-drop}). The performance decrease varied substantially across feature categories: masking laboratory findings led to the largest drop ($-$19.8 percentage points [p.p.]), followed by imaging results ($-$16.6 p.p.) and symptoms ($-$7.2 p.p.) and physical examinations ($-$7.1 p.p.). This pattern is consistent with clinical diagnostic workflows, in which test results are commonly used to refine or revise diagnoses after an initial clinical assessment \cite{mandell_clinical_2024}. These findings motivate the need for follow-up questioning strategies, such as in MedClarify, that actively recover diagnostically relevant information when such features are missing at first presentation.

\subsection*{MedClarify improves diagnostic accuracy under incomplete patient cases}

To evaluate whether follow-up questions generated by MedClarify improve diagnostic accuracy under incomplete patient information, we compared three LLM-based diagnostic systems: (i)~a \textit{single-shot baseline} that predicts diagnoses directly from masked patient cases without follow-up questioning (analogous to the experiments above), (ii)~a \textit{na{\"i}ve multi-turn baseline} that generates follow-up questions heuristically without explicit optimization, and (iii)~\textit{MedClarify}, which selects questions using an information-theoretic criterion. All systems employ the same multi-agent architecture and are evaluated on identical patient cases and masking strategies, so the only difference is their approach to question selection.

\begin{figure}
\centering
\includegraphics[width=0.75\linewidth]{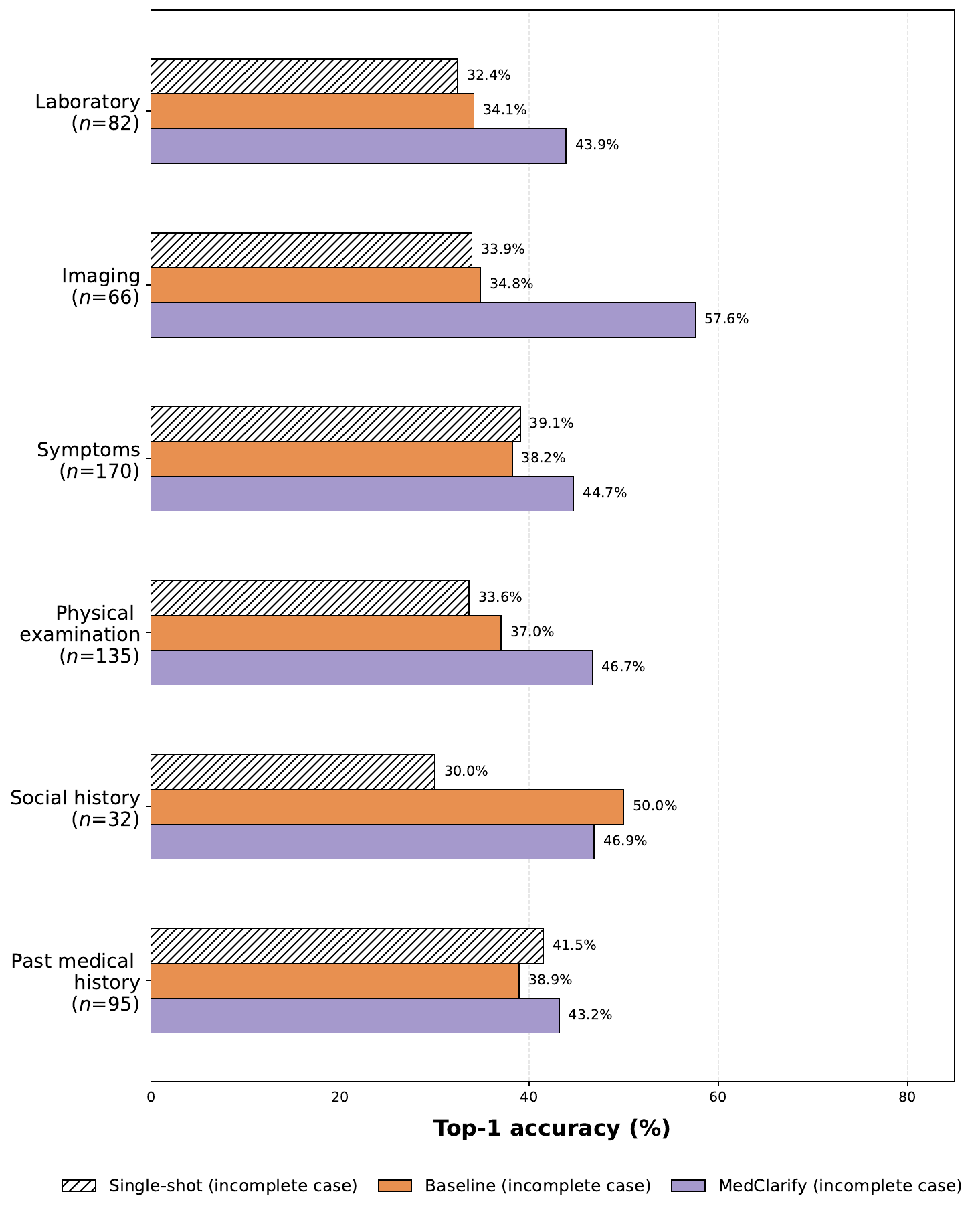}
\caption{\textbf{MedClarify improves diagnostic accuracy under incomplete patient cases.} 
Top-1 diagnostic accuracy is shown for three LLM-based diagnostic systems on the NEJM dataset: a single-shot baseline (striped), a naïve multi-turn baseline with heuristic follow-up questions (orange), and MedClarify with information-theoretic question selection (purple). All LLM-based systems are instantiated with Llama-3.3-70B. Patient cases are evaluated under single-feature masking, where one clinical feature group (laboratory tests, imaging results, symptoms, physical examination, social history, or past medical history) is removed at a time. All systems are evaluated on the same masked patient cases using an identical multi-agent architecture and differ only in their question selection strategy.
}
\label{fig:medclarify-feature-recovery}
\end{figure}

Under incomplete clinical information, MedClarify consistently improved top-1 diagnostic accuracy relative to both baselines across all masked feature categories (Figure~\ref{fig:medclarify-feature-recovery}). When individual feature categories were masked, the na{\"i}ve multi-turn baseline achieved only limited improvement over the single-shot baseline with feature masking, suggesting that the heuristic approach to questions selection is of limited diagnostic utility. In contrast, MedClarify achieved higher top-1 accuracy across all feature categories, with particularly pronounced gains when diagnostically informative features such as laboratory tests and imaging results were missing. Here, when laboratory findings were masked, MedClarify improved top-1 accuracy by 11.5 p.p. over the single-shot baseline and by 9.8 p.p. over the na{\"i}ve multi-turn baseline; imaging masking demonstrated improvements of 23.7 p.p. and 22.8 p.p., respectively. These results indicate that information-theoretic question selection enables effective retrieval of diagnostically relevant information often reaching the potential performance of the evaluation on complete cases. The corresponding results for the top-3 accuracy show consistent trends (Extended Figure~\ref{exfig:top-3-ablation}).

\subsection*{Diagnostic accuracy under joint masking of all clinical features}

We next extend the evaluation to a more challenging setting in which \emph{all} clinical feature categories are masked concurrently. This setting provides a stricter numerical experiments, primarily to better understand when and where information-seeking through follow-up questioning is most effective. Across all three datasets, MedClarify consistently outperformed the na{\"i}ve multi-turn baseline (Figure~\ref{fig:accuracy}a). For example, the top-1 accuracy increased from 28.0\% to 36.1\% (+8.1~p.p.; NEJM dataset), from 41.2\% to 46.1\% (+4.9~p.p.; MediQ); and from 44.3\% to 48.4\% (+4.1~p.p.; MedQA). Aggregated across datasets, MedClarify improved the top-1 diagnostic accuracy by 5.7 p.p. relative to the baseline multi-turn system. For comparison, we also report the improvement relative to a single-shot LLM (+26.9~p.p.; Extended Table~\ref{tab:accuracy}).

Overall, MedClarify required fewer follow-up questions to reach the correct diagnosis in the challenging setting where all clinical feature categories were masked. Specifically, MedClarify averaged 8.96 questions per case compared with 9.81 questions for the naïve multi-turn baseline, indicating more efficient information acquisition through targeted questioning (Extended Figure~\ref{exfig:efficiency-a}). In addition, we evaluated the candidate diagnoses based on the higher Mean Reciprocal Rank (MRR), which measures how early the correct diagnosis appears in the ranked differential diagnosis lis (Extended Figure~\ref{exfig:efficiency-b}). MedClarify consistently achieved a higher MRR than the na{\"i}ve multi-turn baseline across all datasets. On average, MedClarify improved the MRR over the na{\"i}ve baseline by 0.08 (NEJM), 0.05 (MediQ), and 0.05 (MedQA), respectively, which confirms that the correct diagnosis is closer to the top of the ranked list with candidate diagnoses.

\subsection*{Diagnostic accuracy across medical specialties}

\begin{figure}
\centering

% Top row - full width
\llap{\textbf{\textsf{a}}\hspace{0.5em}}%
\includegraphics[width=1\linewidth]{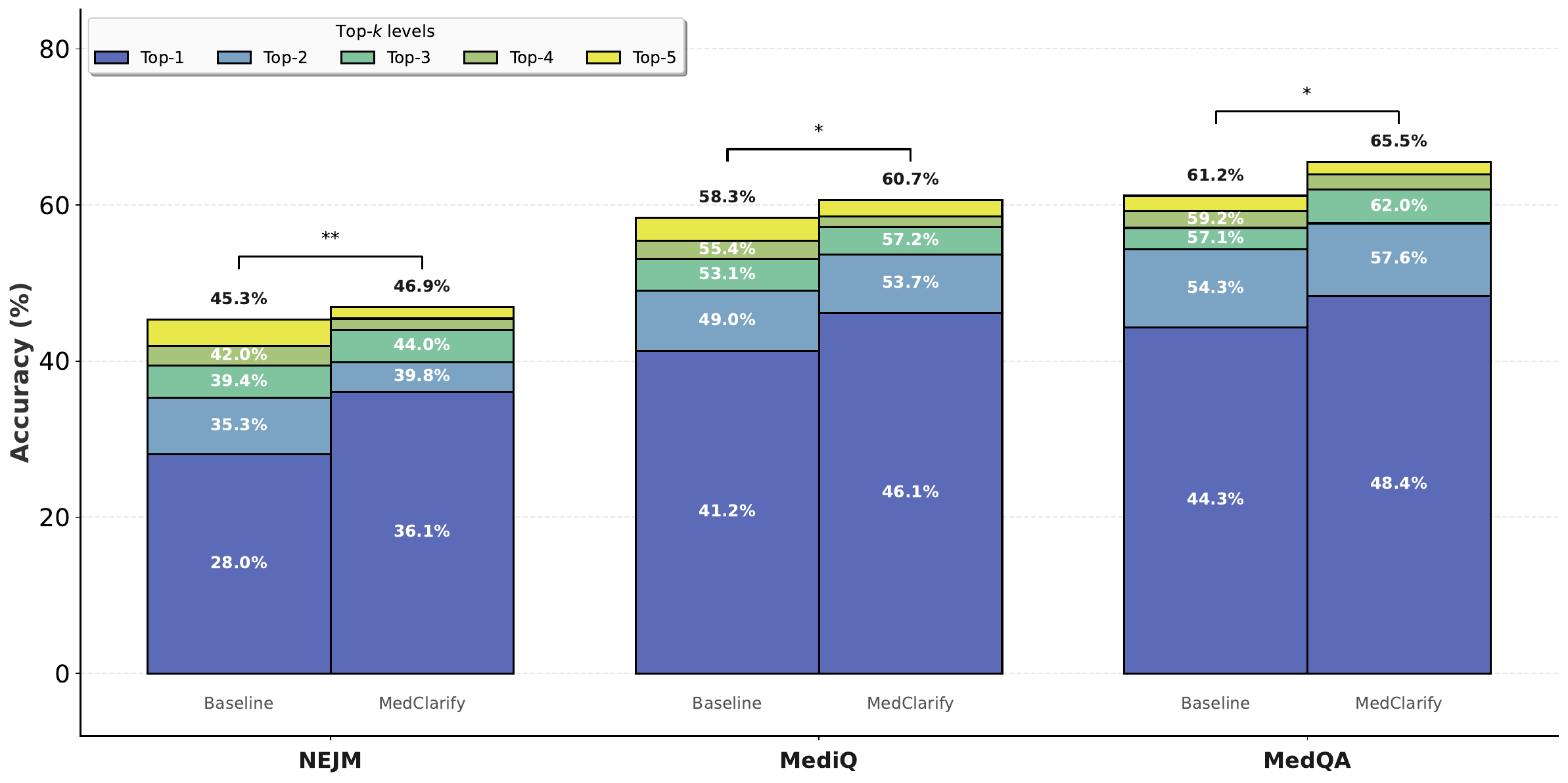}

\vspace{-0.2cm}

% Bottom row - two subfigures side by side
\begin{minipage}[t]{0.48\linewidth}
    \centering
    \llap{\textbf{\textsf{b}}\hspace{0.5em}}%
    \includegraphics[width=\linewidth]{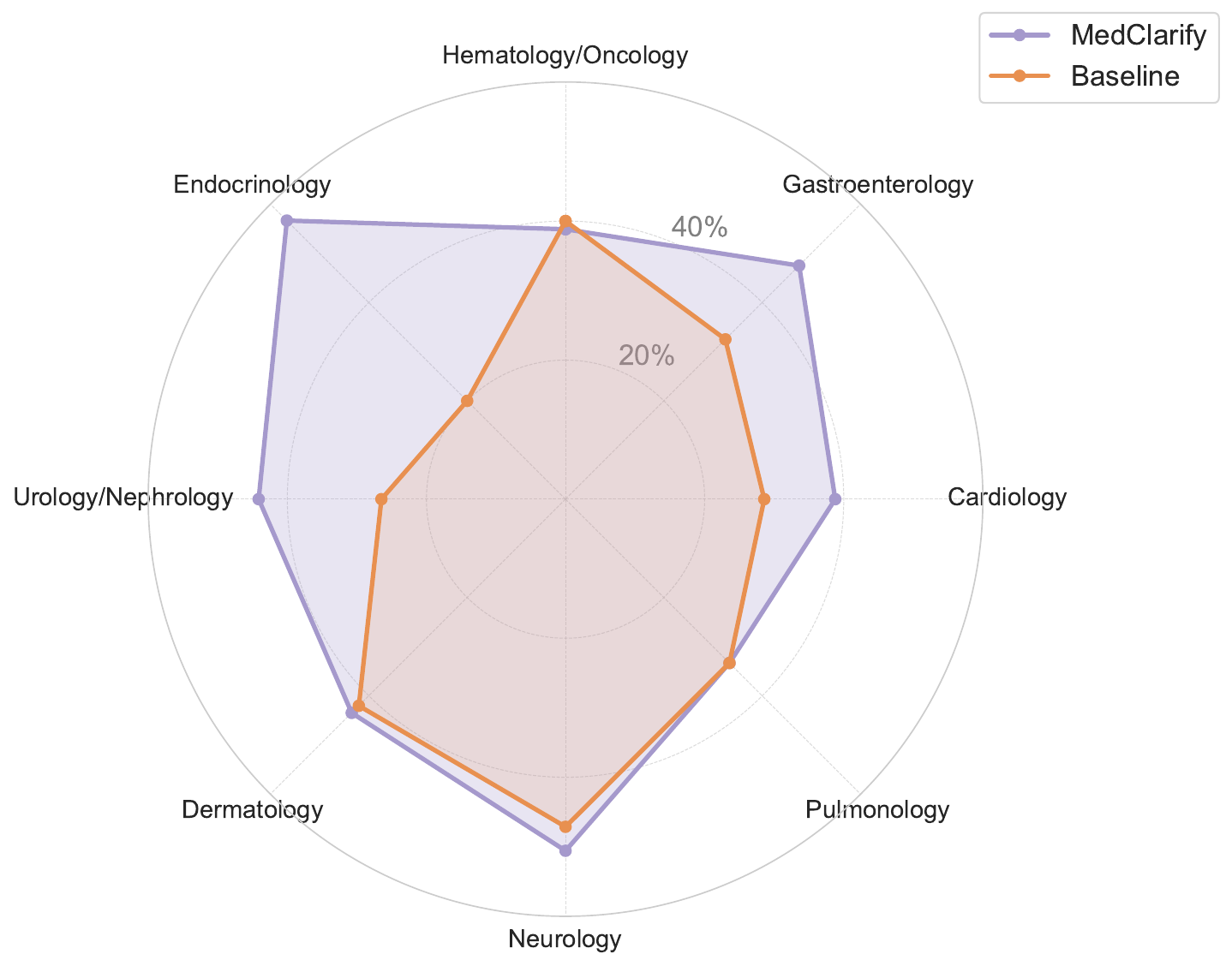}
\end{minipage}
\hfill
\begin{minipage}[t]{0.48\linewidth}
    \centering
    \llap{\textbf{\textsf{c}}\hspace{0.5em}}%
    \includegraphics[width=\linewidth]{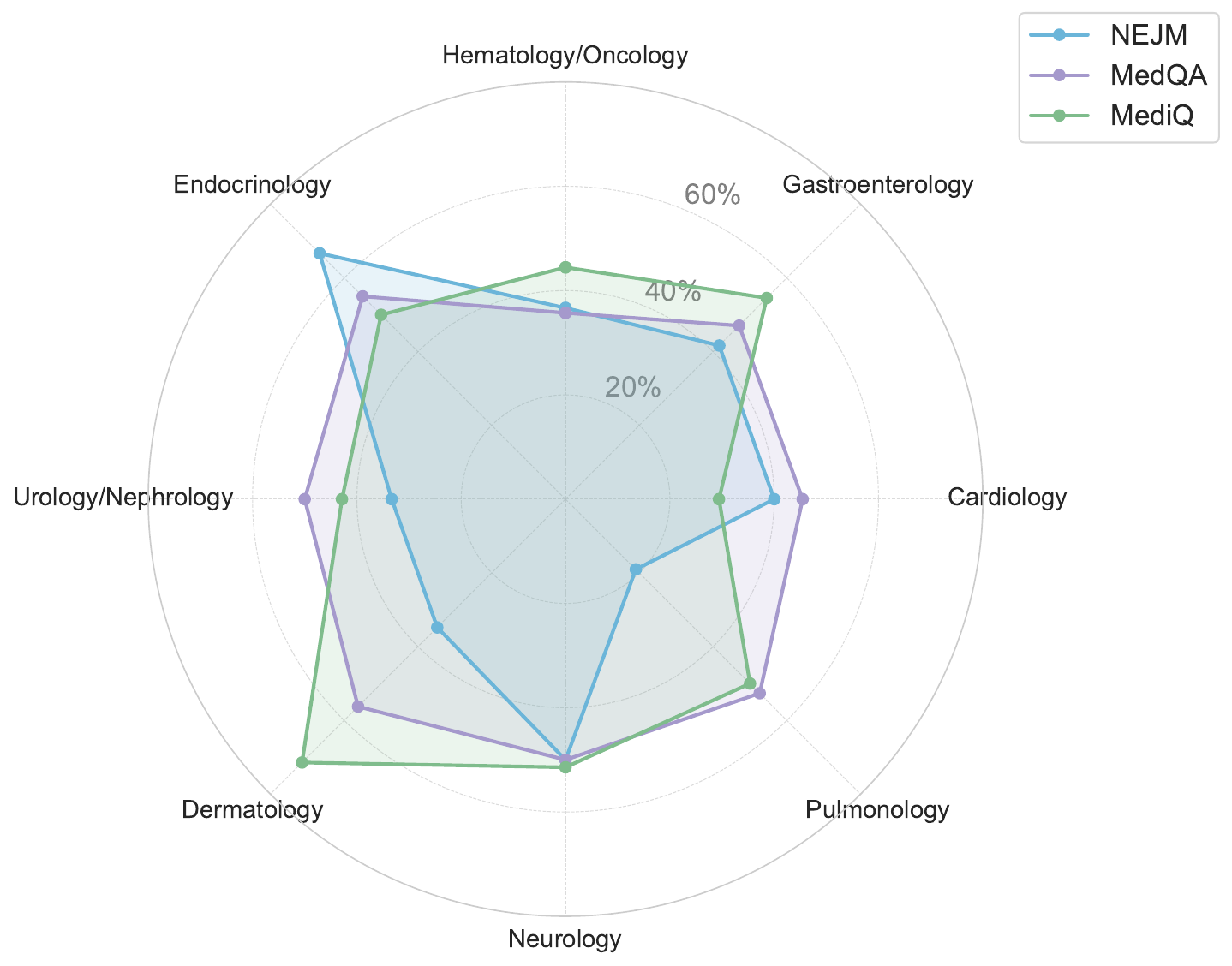}
\end{minipage}

\caption{\textbf{Diagnostic accuracy under joint masking of all feature categories.} The experiments are based on a more challenging setting in which \emph{all} clinical feature categories are masked concurrently, which provides a stricter numerical benchmark to assess when and where information-seeking through follow-up questioning is most effective. \textbf{a},~Comparison of top-$k$ accuracy of MedClarify (instantiated with Llama-3.3-70B as the LLM backbone) and the baseline across the NEJM, MediQ, and MedQA datasets. Statistical significance was tested using 95\% confidence intervals ($^{*}$: $p \le 0.05$; $^{**}$: $p \le 0.01$; $^{***}$: $p \le 0.001$). \textbf{b},~Top-1 diagnostic accuracy of MedClarify (instantiated with Llama-3.3-70B) stratified by medical specialty under joint feature masking, comparing MedClarify with the na{\"i}ve multi-turn baseline. Overall, MedClarify has a strong performance gain across all medical specialties, but with particularly large relative gains for endocrinology. \textbf{c},~Top-1 diagnostic accuracy of MedClarify (instantiated with Llama-3.3-70B) under joint feature masking to demonstrate the overall difficulty of patient cases across datasets.
}
\label{fig:accuracy}
\end{figure}

Diagnostic performance varied across medical specialties  (Figure~\ref{fig:accuracy}b). MedClarify yielded the largest relative gains in endocrinology, where top-1 accuracy increased from 20.0\% to 56.7\% (+36.7~p.p.). The performance of the baseline was comparative low in this specialty, suggesting that structured follow-up questioning is particularly effective in domains where diagnoses depend on targeted clarifications of laboratory-related findings. In contrast, more modest improvements were observed, for example, in dermatology and neurology. In dermatology, accurate diagnosis often depends on visual inspection and pattern recognition from images, which are not captured in text-only patient cases and therefore are challenging for medical LLMs \cite{mijares_validation_2025}. Similarly, neurological diagnoses frequently rely on detailed neurological examinations, imaging, or electrophysiological studies, which are difficult to reconstruct through conversational follow-up alone. More broadly, the performance of MedClarify across medical specialties also reflects the underlying difficulty of the evaluated datasets (Figure~\ref{fig:accuracy}c).

\subsection*{Performance gain is robust across different LLM backbones}

\begin{figure}
\centering

% Top row - full width
\llap{\textbf{\textsf{a}}\hspace{0.5em}}%
\includegraphics[width=.95\linewidth]{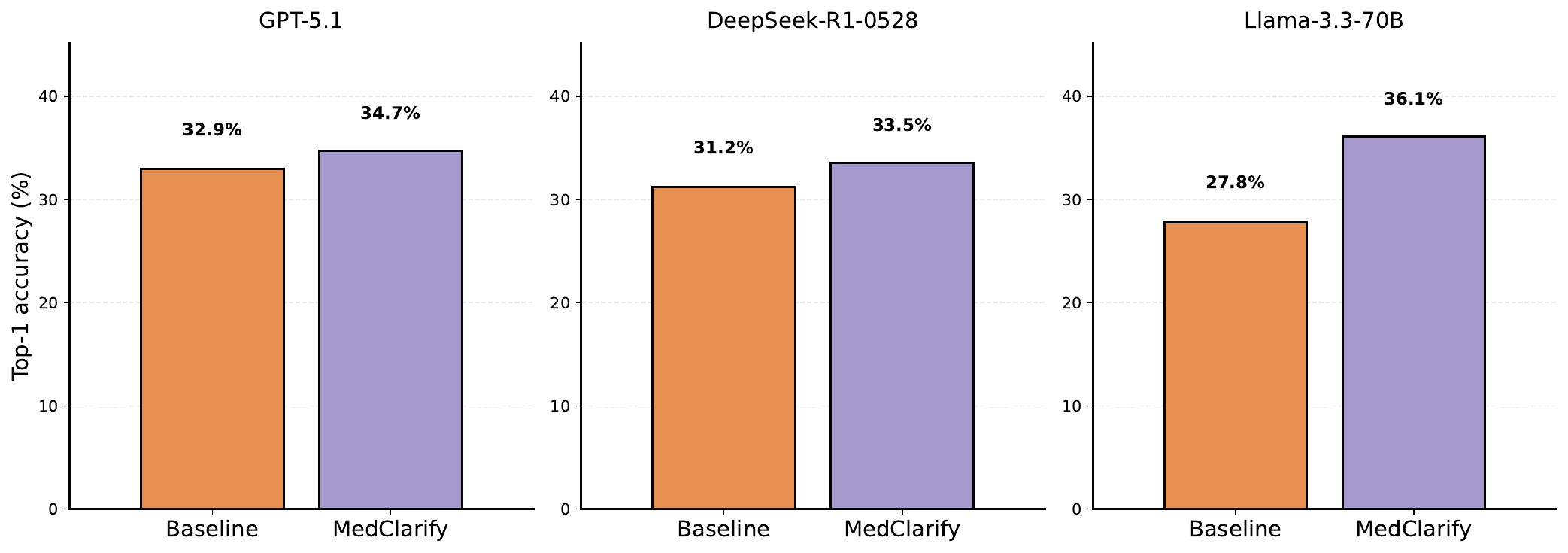}
% \vspace{0.8em}

\llap{\textbf{\textsf{b}}\hspace{0.5em}}%
\includegraphics[width=.95\linewidth]{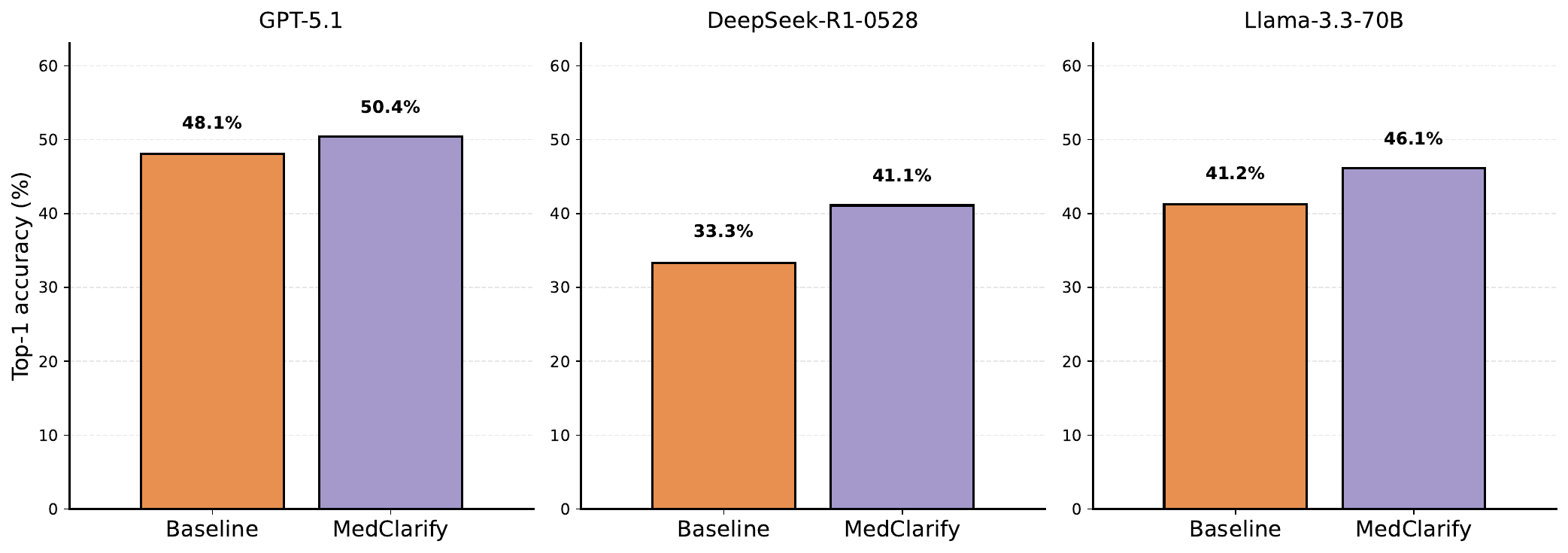}
% \vspace{0.8em}

\llap{\textbf{\textsf{c}}\hspace{0.5em}}%
\includegraphics[width=.95\linewidth]{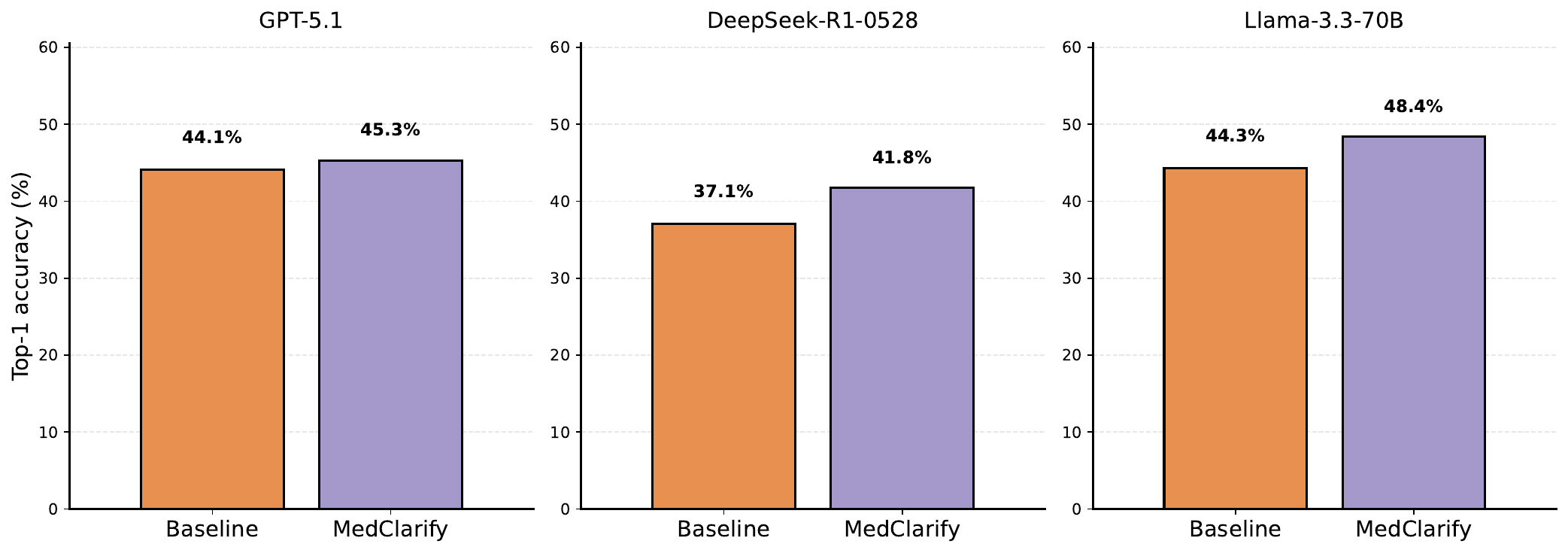}
% \vspace{0.8em}

\caption{\textbf{Overview of the diagnostic accuracy of MedClarify across different LLM backbones.} Here, MedClarify is compared when instantiated with different LLM backbones, namely, GPT-5.1, Deepseek-R1-0528, and Llama-3.3-70B. Reported is the top-1 accuracy of MedClarify (purple) in comparison the na{\"i}ve multi-turn system across different datasets: \textbf{a},~NEJM; \textbf{b},~MediQ; and \textbf{c},~MedQA.}  
\label{fig:backbone}
\end{figure}

We demonstrate the robustness of MedClarify across various LLM backbones. Here, we again draw upon the experiment setting where all feature categories are masked jointly. These results confirm that performance gain from MedClarify is robust across different LLM backbones, including GPT-5.1, Deepseek-R1-0528, and Llama-3.3-70B (Figure~\ref{fig:backbone}). Across the NEJM, MediQ, and MedQA datasets, MedClarify consistently leads to large performance improvements over the na{\"i}ve multi-turn baseline. For instance, on the MediQ dataset, the top-1 accuracy increases from 48.1\% to 50.4\% (+2.3~p.p.; GPT-5.1), 33.3\% to 41.1\% (+7.8~p.p.; Deepseek-R1-0528), and 41.2\% to 46.1\% (+4.9~p.p.; Llama-3.3-70B). The findings remain consistent even when reducing the number of turns in the evaluation (see Extended Figure~\ref{exfig:backbone-iter-five} for results with a maximum of five turns).

\subsection*{Diagnostic belief updating in MedClarify}

\begin{figure}
\begin{center}
\includegraphics[width=1\linewidth]{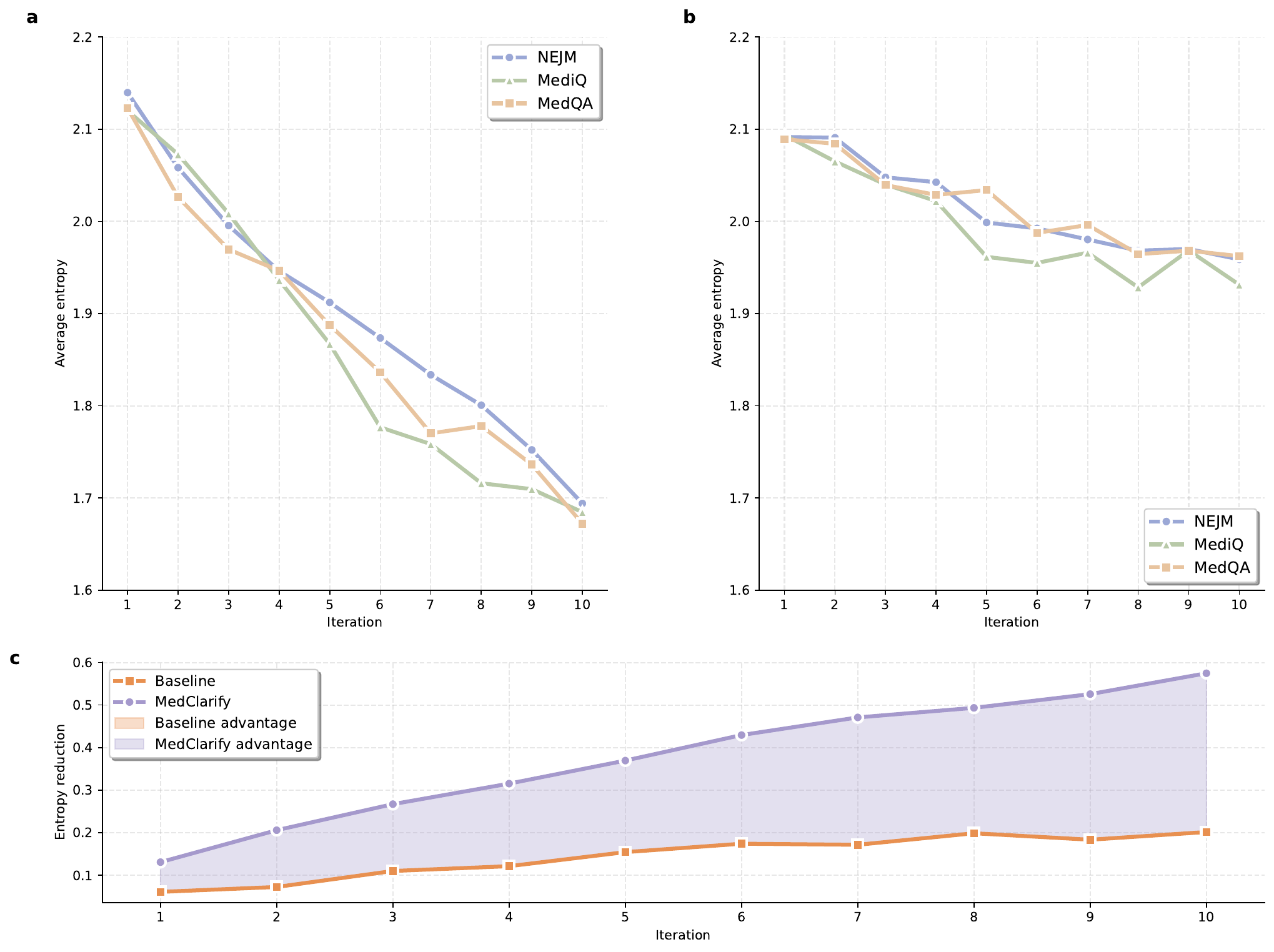}
\end{center}

\caption{\textbf{Analysis of diagnostic belief updating in MedClarify quantified through entropy reduction.} We evaluate Shannon entropy (in bits) of the diagnostic probability distribution across successive turns in the diagnostic dialogue. Entropy measures the concentration of probability mass across candidate diagnoses, with lower values indicating stronger belief concentration. Entropy reduction is computed relative to the initial state (round~0), where all candidate diagnoses are initialized with equal probability. Comparison of the average entropy stratified by dataset (NEJM, MediQ, MedQA) for \textbf{a},~MedClarify vs. \textbf{b},~the na{\"i}ve multi-turn LLM as a baseline. \textbf{c},~Average reduction in entropy per iteration over turns. MedClarify is more effective due to Bayesian belief updating, reflected by a faster concentration of diagnostic beliefs as evidence accumulates. 
}
\label{fig:entropy}
\end{figure}

To examine whether the information-theoretic follow-up questions in MedClarify lead to effective diagnostic belief updating, we measured the reduction in Shannon entropy (in bits) of the diagnostic probability distribution over different turns in the conservation. Here, entropy quantifies the concentration of beliefs across candidate diagnoses, with lower entropy indicating stronger concentration on a smaller set of plausible conditions. We computed the reduction in entropy relative to the initial state (round 0), where all candidate diagnoses are initialized with equal probability (Figure~\ref{fig:entropy}a). Evidently, MedClarify exhibited a consistently larger entropy reduction than the na{\"i}ve mult-turn baseline. This pattern reflects the combined effect of information-theoretic question selection and Bayesian belief updating in MedClarify and indicates that diagnostic beliefs concentrate more rapidly on a single candidate diagnosis as evidence accumulates compared to the na{\"i}ve mult-turn baseline system. The entropy reduction is also consistent across different datasets (Figure~\ref{fig:entropy}b,c).

A key strength of MedClarify is that it produces confidence scores that is clinically meaningful and aligned with diagnostic performance. To demonstrate this, we analyzed how the confidence scores evolve across diagnostic iterations and how well they correspond to the actual diagnostic accuracy (Extended Figure~\ref{exfig:conf}a). Compared to the na{\"i}ve multi-turn baseline, MedClarify yields confidence estimates that are better calibrated with the observed accuracy, which indicates not only more effective belief updating but also that the confidence scores can be informative for decision-making in clinical practice. To quantify the alignment between confidence and accuracy, we further computed the Expected Calibration Error (ECE) \cite{pakdaman_naeini_obtaining_2015}, which measures the discrepancy between predicted confidence and empirical diagnostic accuracy. The naïve multi-turn baseline exhibited a high ECE of 0.37, indicating poor calibration and confidence scores that frequently remain high despite incorrect diagnoses. In contrast, MedClarify achieved an ECE of 0.16, corresponding to a 56.8\% reduction in calibration error relative to the na{\"i}ve multi-turn baseline. Overall, these results indicate that MedClarify produces confidence estimates that more reliably track diagnostic correctness and avoids overconfident predictions compared to the na{\"i}ve baseline.

%%%%%%%%%%%%%%%%%%%%%%%%%%%%%%%%%%%%%%%%%%%%%%%%%%%%%%%%%%%%%%%%%%%%%%%%%%%%%%

\newpage
\section*{Discussion}
\label{sec:discussion}

% context / summary

LLMs are increasingly used for diagnostic decision-making \cite{thirunavukarasu_large_2023, williams_evaluating_2024,clusmann_future_2023,moor_foundation_2023, tu_towards_2025, chen_enhancing_2025, tang_medagents_2024, kanjee_accuracy_2023,eriksen_use_2024, goh_large_2024, hager_evaluation_2024, goh_gpt-4_2025, liu_generalist_2025, olmo_assessing_2024, chen_rarebench_2024, abdullahi_learning_2024,johri_evaluation_2025,wu_medical_2024,seki_assessing_2025,zhuang_learning_2025,takita_systematic_2025,jin_what_2021,gao_leveraging_2025,shieh_assessing_2024,zhou_uncertainty-aware_2025,li_mediq_2024, wang_healthq_2025}, but, in clinical practice, not all required information is available at the initial patient encounter; hence, LLMs rarely have access to all the information needed for an accurate diagnosis. Rather, missing details create diagnostic uncertainty, so physicians routinely ask follow-up questions to rule out competing conditions. To address this, we developed MedClarify, a novel information-seeking AI agent to ask \textit{targeted} follow-up questions. MedClarify offers a paradigm shift by asking proactive questions. For this, MedClarify creates a list of potential candidate diagnoses, computes the probability for each, and then formulates targeted follow-up questions to reduce diagnostic uncertainty. Across several medical datasets, MedClarify increases diagnostic accuracy by 26.9 p.p. compared to a traditional LLM with a single-shot answer and by 5.7 p.p. compared to an LLM with na{\"i}ve follow-up questions, which corresponds to a reduction in diagnostic errors by 32.3\% and by 9.1\%, respectively.

% general critism: LLMs for incomplete cases

Existing evaluations of medical LLMs typically assume that all relevant patient information is fully provided and that each case can be solved uniquely from the case description \cite{kanjee_accuracy_2023,olmo_assessing_2024,chen_rarebench_2024}. This is in sharp contrast to actual clinical practice, where patient information is often incomplete due to various reasons \cite{raven_comparison_2013, han_varieties_2011}: patients may accidentally omit accompanying symptoms, symptom timelines, or comorbidities, while, in other situations, additional examinations or diagnostic tests may be required. As a result, current benchmarks test LLMs only under \emph{idealized} conditions that rarely reflect actual diagnostic workflows. Our experiments show that state-of-the-art LLMs frequently fail for cases with incomplete information, and, as a result, diagnostic accuracy drops by $\sim$29 p.p., corresponding to a relative increase in misdiagnoses by $\sim$54\%. Such a magnitude of diagnostic error is clinically unacceptable and underscores the need for approaches that can handle incomplete patient information. Furthermore, these findings indicate that current evaluation practices overlook an important source of diagnostic error and show the need for more realistic benchmarking. While recent work in medical AI agents moves in this direction \cite{kuhl_human-centered_2025, moor_foundation_2023, zakka_almanac_2024, gatto_follow-up_2025, schmidgall_agentclinic_2024,tang_medagents_2024, fleming_medalign_2024}, standardized benchmarks that reflect the incomplete and iterative nature of real patient encounters are needed to advance the development and evaluation of reliable diagnostic systems.

% innovation of MedClarify

Our work extends prior research on medical LLMs and, more broadly, methods for clinical question answering. Unlike existing medical LLMs, which typically operate in a single-shot without interactive reasoning \cite{johri_evaluation_2025, gatto_follow-up_2025, schmidgall_agentclinic_2024, tang_medagents_2024,jin_pubmedqa_2019,pal_medmcqa_2022,wu_medical_2024,seki_assessing_2025,zhuang_learning_2025,takita_systematic_2025,jin_what_2021,gao_leveraging_2025, shieh_assessing_2024}, our work introduces a principled framework for making diagnostic refinements in a multi-turn setting. For this, we formulate a novel DEIG, which quantifies how much a follow-up question is expected to reduce diagnostic uncertainty with respect to a specific disease. While entropy-based question selection has been explored in natural language processing and question answering \cite{mazzaccara_learning_2024, handa_bayesian_2024, smith_prediction-oriented_2023, rothe_question_2017}, these methods generally ignore the hierarchical structure of medical diagnoses. In contrast, we present a tailored approach that accounts for how closely related different diagnoses are, which allows MedClarify to prioritize follow-up questions that collapse or target entire branches of related conditions. In our experiments, we show that a na{\"i}ve entropy guidance without such adaptation for medical diagnoses is suboptimal, whereas our tailored formulation leads to more effective information seeking and overall better diagnostic accuracy.

% benefits

MedClarify offers several strengths for clinical practice. First, the system is model-agnostic and thus flexible; it can be used with different LLM backbones, including open-source and customized medical LLMs, making it easy to deploy MedClarify across medical institutions with varying computational or regulatory constraints. Second, MedClarify allows for built-in interpretability: at each step, it produces an explicit list of differential diagnoses and computes the associated confidence for each, allowing clinicians to track how the diagnostic reasoning evolves and to intervene or override when needed. As such, MedClarify can be used to support clinicians and agentic applications of LLMs in medical workflows. Third, MedClarify is robust across datasets and medical specialties, which enables seamless integration into clinical workflows even where case formats may differ. Fourth, additional domain knowledge can be integrated in a principled manner through the choice of the prior. Fifth, MedClarify's approach aligns closely with established principles of clinical reasoning taught in medical education. The systematic generation of differential diagnoses, iterative hypothesis testing through targeted questioning, and Bayesian updating of diagnostic probabilities mirror the cognitive processes that physicians are trained to develop. This conceptual alignment offers potential advantages for clinical adoption, as the system's reasoning process may be more intuitive and interpretable to clinicians compared to black-box diagnostic algorithms. Furthermore, MedClarify could serve as an educational tool, demonstrating effective diagnostic questioning strategies to medical trainees and providing a framework for structured clinical reasoning that reinforces best practices in history taking. Together, these strengths make MedClarify a versatile tool for improving the real-world applicability of medical LLMs with the aim of improving diagnostic reasoning capabilities through information-seeking.

% Physical examination

The observed performance differences across medical specialties reflect fundamental limitations of text-based diagnostic systems rather than merely dataset characteristics. Dermatological diagnosis relies heavily on visual pattern recognition---the morphology, distribution, and evolution of skin lesions often cannot be adequately conveyed through verbal description alone. Similarly, neurological diagnosis depends substantially on direct observation and physical examination, including assessment of gait, reflexes, tone, coordination, and subtle findings such as fasciculations or nystagmus that are difficult to capture textually. These limitations suggest that text-based diagnostic LLMs, including MedClarify, which showed poorer accuracy in those medical specialties, may be inherently less suited for specialties where visual and physical examination findings constitute the primary diagnostic evidence. Medical imaging represents a particularly salient example of this challenge: radiological diagnoses depend on the integrated interpretation of multiple imaging parameters (including lesion morphology, signal characteristics, topographic distribution, and dynamic contrast behavior), which are shaped by acquisition protocol and anatomical context in ways that resist reduction to textual description. Incorporating such multi-parametric imaging data into an LLM-based diagnostic framework therefore poses challenges that go beyond simply adding another input modality.

% Med Clarify vs traditional CDSS

MedClarify differs fundamentally from traditional clinical decision support systems (CDSS) in its active, information-seeking approach. Conventional CDSS \cite{sutton_overview_2020, belard_precision_2017, kawamoto_improving_2005} typically operate passively, providing alerts or suggestions based on data already entered into the medical record---for example, flagging potential drug interactions or suggesting diagnoses consistent with documented findings. These systems depend entirely on the completeness and accuracy of information that clinicians choose to document. In contrast, MedClarify actively identifies gaps in diagnostic information and proposes targeted questions to address them, potentially surfacing relevant clinical details that might otherwise remain unelicited. This shift from passive decision support to active diagnostic partnership represents a novel approach that could complement existing CDSS infrastructure. Furthermore, whereas traditional CDSS often suffer from alert fatigue due to excessive, non-specific notifications, MedClarify's information-theoretic approach ensures that suggested questions are targeted and relevant to the specific diagnostic context, potentially improving signal-to-noise ratio and clinician engagement with system recommendations.

% limitations

This study has several limitations that will need to be addressed to further improve agentic LLMs such as MedClarify in medical workflows. First, our evaluation relies on simulated patient interviews, which approximate---but do not fully capture---the variability, ambiguity, and communication patterns encountered in real clinical encounters. In our simulation, we assume that patients provide clear, binary responses to questions, whereas, in reality, patient answers are often ambiguous, uncertain, or qualified. Patients may respond with partial information, express uncertainty about their own symptoms, or provide contradictory statements across the course of an encounter. Furthermore, real patients frequently volunteer unsolicited information that may be diagnostically relevant or, alternatively, distracting and tangential; the constraint that our Patient Agent may reveal only one piece of information per question does not reflect the nature of all clinical conversations. These simplifications may overestimate MedClarify's performance in real-world settings where information extraction from patients is itself a substantial challenge. Second, although MedClarify is model-agnostic, the performance depends on the underlying LLM backbone; frontier models may behave differently as they evolve. As a cautionary note, we emphasize that all LLMs remain susceptible to hallucination or generating fabricated content, which can influence differential diagnosis generation, follow-up questioning, and thus the accuracy of the final diagnosis. Third, our experiments use benchmark datasets that, despite spanning multiple medical specialties \cite{bedi_testing_2025}, may not reflect the full diversity of real-world patient cases, including multimorbidity, atypical presentations, or incomplete histories. Fourth, our approach focuses on textual patient cases and does not yet incorporate additional modalities such as imaging, which are essential for many clinical decisions; future work could extend our framework to these data sources. Fifth, while MedClarify improves information-seeking behavior, it does not evaluate the clinical appropriateness, safety, or downstream consequences of the suggested follow-up questions. As with other medical applications of LLMs, full clinical validation would likely require rigorous evidence from prospective studies or randomized controlled trials. Nevertheless, our experiments provide an important initial step toward developing, evaluating, and comparing LLM-based diagnostic systems through numerical experiments. Together, these limitations indicate that broader validation in real clinical environments and across more diverse datasets is needed to fully assess the effectiveness and safety of medical LLMs systems such as MedClarify in practice.

% broader outlook 

MedClarify demonstrates that information-seeking behavior can strengthen the diagnostic performance of medical LLMs, particularly when patient information is incomplete. As medical AI agents continue to evolve, structured approaches to question selection and uncertainty reduction will likely become central to safe and reliable clinical deployment. Integrating agentic systems such as MedClarify that enable interactive diagnostic dialogues into real clinical workflows offers a promising path toward more accurate AI-assisted diagnosis, improved patient interactions, and reduced misdiagnosis.

\newpage
\section*{Methods}
\label{sec:methods}

\subsection*{Datasets}

% data selection

We evaluated MedClarify using 469 diagnostic cases from three datasets: the NEJM Image Challenge \cite{noauthor_image_nodate}, MediQ \cite{li_mediq_2024}, and MedQA \cite{jin_pubmedqa_2019}. We included only cases that (i) focus on diagnostic reasoning (i.e., rather than general questions aimed at probing clinical knowledge such as ``What is the most likely organism?'') and (ii) provide sufficient clinical detail to simulate a doctor–patient interaction (i.e., cases with sufficient length and detail). We thus excluded cases that focus on general disease management, treatment, prognosis, or other non-diagnostic tasks. From the 400 cases published between September 2017 to May 2025 as part of the NEJM Image Challenge, 170 met these criteria. From the 1,272 and 1,272 cases in the MediQ and MedQA datasets, 129 and 170 cases were included, respectively. The identifiers for the cases are provided in the code repository. Overall, the cases span eight medical specialties: cardiology, pulmonology, gastroenterology, neurology, dermatology, endocrinology, hematology/oncology, and urology/nephrology (Extended Figure~\ref{fig:dist}).

% preprocessing

We preprocessed the patient cases to avoid confounding due to differences in formatting and to be able to analyze the effect of missing clinical information. For this, we converted each case into a structured case arranged in a JSON format using DeepSeek-R1-0528. Each case is organized into six predefined medical feature categories: symptoms, social history, past medical history,  physical examination, laboratory tests, and imaging results. This structure allows us to use feature masking in our experiments later to assess the effect of different features on the overall diagnostic accuracy while ensuring otherwise comparable input across models and datasets. Importantly, not all cases contain all six features  (e.g., social history is frequently missing), and, if so, the corresponding features are then simply left empty. Summary statistics are in Extended Table~\ref{tab:medical-feature}. An example of a structured case is given in Extended Listing~\ref{exlst:structured-patient-case}.

\subsection*{MedClarify for information-seeking in medical AI agents}

% overview

MedClarify performs multi-turn diagnostic refinement by taking an initial patient case as input and then producing differential diagnoses (i.e., a set of candidate diagnoses) that are iteratively updated as new information becomes available. In each turn $t$, MedClarify asks a clarification question, and, based on the patient’s response, updates the internal belief state to make a refined diagnosis (Figure~\ref{fig:flowchart}a). 

Methodologically, MedClarify operates in four steps: (1)~computing a candidate list of diagnoses with corresponding probabilities, where a broad probability distribution reflects diagnostic uncertainty (\textit{assessment of candidate diagnoses}); (2)~generating candidate follow-up questions designed to refute individual diagnoses or to explore previously unconsidered conditions (\textit{question generation}); (3)~selecting the follow-up question expected to reduce that uncertainty using DEIG (\textit{question selection}); (4) updating the probability distribution through a Bayesian framework, so that the posterior distribution reflects the newly gathered evidence (\textit{Bayesian updating}). These steps are repeated until the system reaches sufficient confidence in a single diagnosis (Figure~\ref{fig:flowchart}b). Each step is detailed in the following.

In MedClarify, LLMs are used at two points in the above process: in Step~1 to compute the candidate list of diagnoses with associated probabilities, in Step~2 to generate candidate follow-up questions, and in Step~3 to select the highest DEIG question among the candidate questions. Importantly, the choice of LLM backbone is flexible, and we later evaluate MedClarify using several state-of-the-art models.

\noindent
\underline{Step 1: Assessment of candidate diagnoses}

MedClarify takes the current patient case $C_t$ during turn $t$ as input and then produces a differential diagnosis represented as a set of candidate diseases with confidence score, i.e., $D_t = \{(d_i, \pi_i)\}_{i=1}^{k}$, where $\pi_i \in [0, 1]$ denotes a confidence score that disease $d_i$ is the true diagnosis. The candidates are sorted by confidence score (i.e., $\pi_1 \geq \pi_2 \geq \ldots$), so that $d_1$ is the top-ranked diagnosis at turn $t$.

To compute the differential diagnoses, an LLM is prompted with the patient case $C_t$. Here, we instruct the LLM to produce a differential diagnosis using a JSON schema with $k$ candidate diseases, each accompanied by a confidence score $\pi_i$ indicating how strongly the model believes that disease $d_i$ matches the true underlying condition. As we show in our experiments, it is sufficient that the confidence scores provide only a meaningful ranking of candidate diagnoses; the scores do \textit{not} need to be well-calibrated probabilities. This flexibility allows MedClarify to operate with both general-purpose and specialized medical LLMs, many of which do not produce calibrated probability estimates. Nevertheless, if desired, the framework could be extended to incorporate calibrated probabilities \cite{farquhar_detecting_2024, nakkiran_trained_2025, hekler_test_2023}; in our case, we use temperature scaling \cite{guo_calibration_2017, yu_robust_2022} for this purpose. 

The prompt is stated in the following. Therein, we make use of chain-of-thought reasoning \cite{wei_chain--thought_2022}, which has been shown to improve the reasoning accuracy of medical LLMs \cite{savage_diagnostic_2024, zhou_large_2025}. The confidence score is generated with a lower bound of 0.1 instead of 0.0 to include only plausible diagnoses.

\begin{tcolorbox}[
    colback=gray!10,
    colframe=gray!150,
    title={\textbf{Prompt for generating differential diagnosis}},
    fonttitle=\bfseries,
    arc=2mm,
    boxrule=1.5pt,
    left=20pt,
    right=15pt,
]
\singlespacing
\vspace{-1cm}
\footnotesize
\begin{lstlisting}[language=json]
Your goal is to give a differential diagnosis and assign each a confidence score [0.1 - 1] based on <Task> and further information from <Inquiries History> if there is.
Your diagnoses should be a specific disease name, not a general diagnosis. Try to rule out diseases in your previous <Differential Diagnosis> if possible.
Output a value of 0.9-1.0 if the disease is expected to be the correct diagnosis; 0.6-0.9 if it is a possible diagnosis; and below 0.6 if it is unlikely.

Here are some examples:
<Peripheral neuropathy> -> too general, <Guillain-Barre syndrome> -> specific
<Hemolytic anemia> --> too general, <Sickle Cell Disease> --> specific
\end{lstlisting}

\end{tcolorbox}

In our experiments, we use $k=5$ differential diagnoses. Our choice for $k=5$ differential diagnoses was guided by clinical practice to offer a reasonable tradeoff between clinical practicality for physicians and potential gains in diagnostic accuracy. In particular, this choice was previously validated in medical LLM  \cite{mcduff_towards_2025,spitzer_effect_2025} where it gives a good empirical performance, but where large values lead to only marginal incremental gains. Further, Step~3 requires $2 \times k$ prompts, so the choice of $k=5$ offers a good trade-off in terms of computational scalability.

\noindent
\underline{Step 2: Question generation}

In the question generation step, MedClarify uses the current set of candidate diagnoses $D_t$ to produce a set of candidate follow-up questions $Q_t = \{q_1, \ldots, q_n\}$. For this, an LLM is prompted in two ways: (i)~to generate questions $q_1, \ldots, q_{n-1}$ for which the answers will help refute or support a specific diagnostic alternative, and (ii)~to generate an exploratory question $q_n$ to account for conditions that are not considered in $D_t$. 

For (i), the idea is to generate `discriminatory' questions that primarily compare the most likely disease $d_1$ against potential alternatives $d_2, d_3, \ldots$ Hence, we design questions that compare the top-ranked disease against the different diagnoses as follows: 
\begin{itemize}
\item \emph{Question 1 ($q_1$)} differentiates between the top-1 disease ($d_1$) and the top-2 disease ($d_2$), aiming to refute $d_1$ and support $d_2$;  
\item \emph{Question 2 ($q_2$)} differentiates between the top-1 disease ($d_1$) and the top-3 disease ($d_3$), aiming to refute $d_1$ and support $d_3$; etc. 
\end{itemize}
Here, we make again use of chain-of-thought reasoning \cite{wei_chain--thought_2022} to guide the model to carefully reflect the available evidence. The corresponding prompt is as follows:

\begin{tcolorbox}[
    colback=gray!10,
    colframe=gray!150,
    title={\textbf{Prompt for generating confirmatory or refuting questions}},
    fonttitle=\bfseries,
    arc=2mm,
    boxrule=1.5pt,
    left=20pt,
    right=15pt,
]
\singlespacing
\vspace{-1cm}
\footnotesize
\begin{lstlisting}
Generate a question to help eliminate <Disease A> but confirm <Disease B> for <Patient Case>. 
The question should be bite-sized and not more than 1 sentence. Do not repeat the questions you have asked before in <Inquiries History> and <Questions>.
You are not allowed to reveal explicitly your diagnosis that you guessed.
<Patient Case>: {}
<Disease A>: {}
<Disease B>: {}
<Inquiries History>: {}
Output your answer concisely in the following format:

Thought:
[Your thought process here]

Response:
[Your question here]
\end{lstlisting}

\end{tcolorbox}

For (ii), we prompt the LLM to explore alternatives that go beyond the existing diagnoses $d_1, \ldots, d_k$. Here, we also make use of chain-of-thought reasoning \cite{wei_chain--thought_2022} to make the reasoning process transparent and to guide the model in carefully comparing potential candidates beyond the currently proposed candidate diagnoses.

\begin{tcolorbox}[
    colback=gray!10,
    colframe=gray!150,
    title={\textbf{Prompt for generating exploratory questions}},
    fonttitle=\bfseries,
    arc=2mm,
    boxrule=1.5pt,
    left=20pt,
    right=15pt,
]
\singlespacing
\vspace{-1cm}
\footnotesize
\begin{lstlisting}
Generate a question to discover other possible diseases apart from <Current Diagnosis> for <Patient Case>.
The question should be bite-sized and not more than 1 sentence. Do not repeat the questions you have asked before in <Inquiries History> and <Questions>.
You are not allowed to reveal explicitly your diagnosis that you guessed.
<Patient Case>: {}
<Current Diagnosis>: {}
<Inquiries History>: {}
Output your answer concisely in the following format:

Thought:
[Your thought process here]

Response:
[Your question here]
\end{lstlisting}

\end{tcolorbox}

\noindent
\underline{Step 3: Question selection}

%\TODO{HUI: How do you compute P(Dt | support) vs P(Dt | refute) -- are just setitng the confidence score to zero? Or do you prompt the LLM saying what would be your distribution *if* question is supported vs  not? --hm: Yes, I prompted LLM to predict the distribution for supported response (answer 'yes' to the question) and refuted response (answer 'no'to the question). That's why the inference time for my method takes so long, because I have 5*2 distribution to simulate for 10-iteration questions. }

MedClarify selects the question $q^\ast$ that is expected to reduce diagnostic uncertainty the most. Formally, MedClarify evaluates each candidate question $q \in Q_t$ using our so-called diagnostic expected information gain (DEIG). The DEIG is an information-theoretic score that quantifies how informative the question is with respect to the current differential diagnosis. For this, MedClarify simulates how the diagnostic distribution would change under two hypothetical responses to the question. For each candidate question, the system queries the LLM twice:
\begin{enumerate}[label=(\alph*)]
\item \emph{Confirmatory case $Diag^+$:} the LLM is prompted under the assumption that the patient answers ``yes'' (i.e., the information supports the targeted diagnosis). Under this assumption, the LLM generates a new differential diagnosis $\mathit{Diag}^+$ and an updated posterior distribution $P(D_{t+1} \mid \mathrm{support})$.
\item \emph{Refute case $Diag^-$:}  the LLM is prompted under the assumption that the patient answers ``no'' (i.e., the information refutes the targeted diagnosis). Under this assumption, the LLM produces a new differential diagnosis $\mathit{Diag}^-$ and a posterior distribution $P(D_{t+1} \mid \mathrm{refute})$.
\end{enumerate}
The two simulated posteriors reflect how the diagnostic landscape would evolve depending on the patient’s answer. The DEIG then quantifies how much diagnostic uncertainty is expected to decrease across these two outcomes. The question with the highest DEIG is selected and posed to the patient agent in the interactive diagnostic loop. This allows the system to rule out competing hypotheses or explore new diagnostic possibilities in a principled manner.

\vspace{0.2cm}

\noindent
\emph{Diagnostic expected information gain (DEIG)}

At a high level, DEIG builds on classical entropy-based information gain, which is a well-established concept in information theory and machine learning, and is often used in question answering and active learning \cite{smith_prediction-oriented_2023, hu_uncertainty_2024, lindley_measure_1956}. However, MedClarify extends this principle in two main ways: (i)~by incorporating how related disease codes are, which allows MedClarify to prioritize questions that rule out entire branches of related conditions; and (ii)~by introducing a regularization term that encourages sufficient exploration, thereby ensuring broader diagnostic coverage and reducing the risk of overlooking rare conditions. Formally, the optimal question under the DEIG is given by 
\begin{equation}
    q\ast = \arg\max_{q \ in Q_t} \left[  \alpha \cdot \mathrm{IG}(q) + \beta \cdot \mathrm{Div}(q) + \gamma \cdot \mathrm{Con}(q) \right] ,
\end{equation}
where $\mathrm{IG}$ is the standard information gain,  $\mathrm{Div}$ is a divergence term that measures the semantic similarity between ICD codes, and $\mathrm{Con}$ quantifies the `concentration' in the distribution of the confidence scores to balance exploration, with parameters $\alpha, \beta, \gamma$. Intuitively, the latter term avoids potential confirmation bias during information-seeking by also explicitly considering potential diagnoses that are not yet among the selected candidates. We specify the different components in the following:

\begin{itemize}
    \item \textbf{Information gain (IG)} follows the standard entropy-based definition \cite{shannon_mathematical_1948, cover_entropy_2005}. In our medical setting, it quantifies how much a candidate question is expected to reduce diagnostic uncertainty. Let $H$ define the (statistical) entropy given by $H(X) = - \sum_{x\in \mathcal{X}} p(x) \log p(x)$. Accordingly, we compute the entropy over the distribution of the confidence scores in both hypothetical cases (a) and (b); that is, $H(P(D_t \mid \mathrm{support}))$ denotes the entropy under supporting evidence, and $H(P(D_t \mid \mathrm{refute}))$ denotes the entropy under refuting evidence. Let $P(D_t)$ denote the distribution of the confidence scores before asking the question. Then, the IG is defined by 
    \begin{equation}
        \mathrm{IG} = H(P(D_t))- \left[\frac{1}{2} \cdot H(P(D_t \mid \mathrm{support})) + \frac{1}{2} \cdot H(P(D_t \mid \mathrm{refute})) \right] . 
    \end{equation}
    The information gain is positive when entropy decreases, meaning the confidence scores become more concentrated around fewer diseases. MedClarify prioritizes questions with high IG because these are expected to produce the greatest reduction in diagnostic uncertainty, i.e., by making the distribution of the confidence scores concentrated on only a few likely diagnoses. 
    
    \item \textbf{Relatedness of ICD codes (Div).} This is a divergence term that captures how different the diagnoses are under the two hypothetical outcomes of a clinical question (support vs. refute). This term favors questions for which the answers lead to very different diagnoses (that is, diagnoses that are less related), so that, essentially, questions are favored that can also rule out related conditions. It is defined as
    \begin{equation}
        \mathrm{Div} = 1 - \mathrm{similarity}(\mathit{Diag}^+,\mathit{Diag}^-)
    \end{equation}
    where $\mathit{Diag}^+$ and $\mathit{Diag}^- $, respectively, represent the sets of candidate diagnoses for supporting and refuting responses to the simulated question. Details about the similarity computation are provided later. 
    
    \item \textbf{Concentration around (Con).} This term measures how `concentrated' the candidate diagnoses are and thus allows to encourage answers that promote broader exploration. We quantify the concentration using the Gini coefficient over the hypothetical outcomes, i.e., 
    \begin{equation}
        \mathrm{Con} = \frac{1}{2} \left[ (1 - \mathrm{Gini}(\mathit{Diag}^+ )) + (1 - \mathrm{Gini}(\mathit{Diag}^- )) \right] ,
    \end{equation}
    where the Gini coefficient is defined as
    \begin{equation}
         \mathrm{Gini} = \frac{\sum_{k=1}^n (2k-(n+1)) x_k }{n \sum_{k=1}^n x_k} .
     \end{equation}
    The Gini coefficient ranges from 0 (i.e., perfectly uniform distribution, meaning that all diagnoses are equally likely) to 1 (i.e., maximum concentration, meaning that the diagnoses are concentrated on a single condition). Lower Gini values therefore indicate that diagnoses span multiple medical areas, which helps MedClarify to encourage exploration where needed to support a more comprehensive diagnostic search that also accounts for rare conditions.
\end{itemize}
For calculating Div and Con, the MedClarify system uses a medical terminology lookup that maps the predicted diseases to standardized disease codes based on ICD-11 \cite{drosler_icd-11_2021}, and then measures how related the conditions are based on the chapter categories. Importantly, this mapping also ensures that potential synonyms for conditions are mapped onto standardized disease names. Here, we use the ICD-11 search API \cite{drosler_icd-11_2021} for retrieving the disease codes. For estimating ICD hierarchical relationships based on their chapter categories, we ask LLMs to generate approximations of ICD chapter similarities (see Table~\ref{tab:similarity_matrix}). The generated similarities are pre-recorded for computational efficiency. The similarity computation is as follows:
\begin{equation}
    \text{similarity}(Diag^+, Diag^-) = 
    \frac{1}{\lvert Diag^+\rvert \times \lvert Diag^-\rvert}
    \sum_{d_i^+ \in Diag^+} \sum_{d_j^- \in Diag^-}
    \operatorname{sim}\bigl(\operatorname{chapter}(d_i^+), \operatorname{chapter}(d_j^-)\bigr) ,
\end{equation}
which simply adds up the pre-recorded similarity for every pair ($d_i^+$,$d_j^-$) and takes its mean, where $Diag^+ = \{d_1^+, \ldots, d_i^+\}$ and $Diag^- = \{d_1^-, \ldots , d_j^-\}$. The example of the operation of $sim(\cdot)$ is as follows: $sim(c_1,c_5) = 0.20$, $sim(c_3, c_2) = 0.50$, etc.

\begin{table}[htbp]
    \centering
    \caption{LLM-generated similarity matrix for example of ICD chapters 1--5 which enter MedClarify as domain knowledge.}
    \label{tab:similarity_matrix}
    \begin{tabular}{l|ccccc}
        \hline
        Chapter ($c_i$) & $c_1$ & $c_2$ & $c_3$ & $c_4$ & $c_5$ \\
        \hline
        $c_1$ & 1.00 & 0.35 & 0.35 & 0.65 & 0.20 \\
        $c_2$ & 0.35 & 1.00 & 0.50 & 0.40 & 0.25 \\
        $c_3$ & 0.35 & 0.50 & 1.00 & 0.60 & 0.30 \\
        $c_4$ & 0.65 & 0.40 & 0.60 & 1.00 & 0.40 \\
        $c_5$ & 0.20 & 0.25 & 0.30 & 0.40 & 1.00 \\
        \hline
    \end{tabular}
\end{table}  

\noindent
\underline{Step 4: Bayesian updating} 

After receiving a new patient response at turn $t$, MedClarify updates the diagnostic belief state by computing a posterior distribution over the candidate diagnoses. This update follows Bayes’ rule \cite{smith_prediction-oriented_2023,handa_bayesian_2024}, allowing the system to incorporate new evidence $e$ while retaining information from previous turns. Formally, 
\begin{equation}
    P(d \,\mid\, e)\propto\ P(e \,\mid\, d)\cdot P(d) ,
\end{equation}
where $P\left(d\right)$ is the prior, representing the current belief about the differential diagnosis before incorporating the new evidence; $P(e \,\mid\, d)$ is the likelihood, quantifying how probable the newly obtained evidence is under each candidate in differential diagnosis; and $P(d \,\mid\, e)$ is the posterior, representing the updated confidence for each diagnosis after integrating the prior belief with the new evidence.

One might expect that simply appending new information to the case description of a patient would be sufficient for refining the differential diagnosis, since the LLM receives a more complete case description at each turn. However, updating the case alone does not guarantee that the confidence scores evolve in an evidence-consistent manner where the confidence scores reflect the probability of each diagnosis. The reason is that, because the confidence scores are generated independently at every turn, LLMs tend to anchor on the same few salient parts of the case description; as a result, a na{\"i}ve approach may disregard previously inferred diagnostic information. This can lead to inconsistent diagnosis rankings across turns and hinders how support or refuting evidence is accumulated. By contrast, Bayesian updating explicitly integrates the new evidence with the existing diagnostic distribution, ensuring that the belief state reflects both the updated patient case and the updated probability of each disease, given the past patient dialogue. Further, the patient case is updated with the new evidence, which is later controlled by the \textsc{Update} agent.

To avoid penalizing newly introduced diagnoses at turn $t$, we assign their confidence score $\pi_i$ (i.e., prior probability) as the average prior over the existing diagnoses rather than setting it to zero or an arbitrarily small value. After the Bayesian update, we apply temperature scaling to smooth the posterior distribution and prevent overly peaked probability assignments. Temperature scaling \cite{guo_calibration_2017, yu_robust_2022} is defined as
\begin{equation}
 P_i = \frac{e^{z_i/T}}{\sum_{j} e^{z_j/T}} ,
\end{equation}
where $z_i$ denotes the unnormalized log-probability for diagnosis $d_i$ and $T > 0$ is the temperature parameter.

After each turn, the system evaluates whether to terminate the diagnostic refinement. The process stops when one of the following criteria is met: the maximum number of turns is reached (here: $t=10$), the top-ranked diagnosis exceeds a confidence threshold (here: $P_{\max} > 0.97$), or the confidence gap between the top-1 and top-2 diagnoses exceeds a defined threshold ($\Delta P_{\max} > 0.85$). When the multi-turn process is terminated, MedClarify returns a differential diagnosis consisting of the top five diseases along with their calibrated confidence scores.

\subsubsection*{Implementation details}

In our main experiments, we set the weights of the three DEIG components to $\alpha = 0.5$ (information gain), $\beta = 0.35$ (semantic discrimination), and $\gamma = 0.15$ (diagnostic breadth). These values were chosen to balance reducing diagnostic uncertainty, separating competing differential diagnoses, and maintaining diagnostic breadth. We further conducted a sensitivity analysis by varying each weight while holding the others fixed, and found that the overall performance of MedClarify was largely robust to these choices. After each Bayesian update, we apply temperature scaling with $T=1.1$ to smooth the posterior distribution and prevent the probability of the top-ranked diagnosis from becoming overly peaked, thereby delaying termination based on the stopping criterion ($P_{\max} > 0.97$).

The underlying LLM in our system uses an open-source model from Meta’s Llama family, specifically Llama-3.3-70B-Instruct-Turbo, hosted by DeepInfra. In the LLM configuration, the temperature was set to 0.3, minimum $p$ to 0.1, top-$p$ to 0.9, the repetition penalty to 0.9, and the random seed to 42.

\subsection*{Agentic evaluation framework}

We evaluate MedClarify using an agentic evaluation framework that simulates multiple turns of doctor–patient interactions. The evaluate is based on four LLM-based agents: (i)~a \textsc{Patient} agent that responds to questions; (ii)~a \textsc{Doctor} agent that performs diagnostic reasoning with MedClarify, generating differential diagnoses and follow-up questions; (iii)~a \textsc{Update} agent that integrates the newly retrieved information into the patient case; and (iv)~an \textsc{Evaluator} agent that assesses whether the final diagnosis is correct while accounting for variations in medical terminology. Together, these components create a controlled setting for evaluating how well MedClarify gathers new evidence from patients' responses and improves diagnostic accuracy as a result.

\noindent
\underline{\textsc{Patient} agent} 

The \textsc{Patient} agent simulates the patient and is provided with the \textit{full} case description (i.e., including the clinical details that may be masked in the experiments). The agent is instructed to answer the questions from the \textsc{Doctor} agent strictly based on factual information from the case description. To ensure fair and controlled evaluations, we impose several constraints: the \textsc{Patient} agent may reveal only one piece of information per question, must not mention disease names explicitly (i.e., so that the solution is not revealed), and may not provide speculative or interpretive statements. The following prompt is used:

\begin{tcolorbox}[
    colback=gray!10,
    colframe=gray!150,
    title={\textbf{\textsc{Patient} agent (to answer clinical question)}},
    fonttitle=\bfseries,
    arc=2mm,
    boxrule=1.5pt,
    left=20pt,
    right=15pt,
]
\singlespacing
\vspace{-1cm}
\footnotesize
\begin{lstlisting}[language=json, xleftmargin=0pt]
You are a patient who only answers the Doctor's <Question> based on your given conditions <Task>.
Things you must NOT do:
- Reveal your disease explicitly
- Give an answer that is not the fact based on your given conditions <Task>
- Have any bias in your answer

Here is an example:
**Example task**: She was a lifetime nonsmoker and reported no fevers, joint aches, eye pain, or rashes.

Doctor: "Do you smoke or have a history of smoking?"
Response: "No, I've never smoked."

Doctor: "Any recent fevers or chills?"
Response: "No fevers or chills."

Doctor: "Are you having any joint pain or stiffness?"
Response: "No joint aches or pain."

Doctor: "Any recent weight loss or night sweats?"
Response: "There is no information mentioned about weight loss or night sweats."

Output your answer concisely in the following format:

Response:
[Your answer here]
\end{lstlisting}

\end{tcolorbox}

\noindent
\underline{\textsc{Doctor} agent}

The \textsc{Doctor} agent triggers the MedClarify framework and executes one turn of the diagnostic process. As such, it generates differential diagnoses, generates candidate questions, and selects a follow-up question (using the MedClarify framework from above). After receiving the response from the \textsc{Patient} agent, the \textsc{Doctor} agent also updates the diagnostic probabilities through the Bayesian updating step inside MedClarify. When the termination condition is met, the \textsc{Doctor} agent outputs the final differential diagnosis for evaluation (i.e., via the \textsc{Evaluator} agent).

\noindent
\underline{\textsc{Update} agent}

The \textsc{Update} agent appends the newly retrieved information to the patient case. For this, each question–answer pair is converted into a single concise sentence and appended to the case description. During this process, the agent filters out irrelevant, misleading, or non-informative responses (e.g., ``I am not sure'') to prevent noise that could interfere with diagnostic reasoning. The following prompt is used:

\begin{tcolorbox}[
    colback=gray!10,
    colframe=gray!150,
    title={\textbf{\textsc{Update} agent (to update patient cases)}},
    fonttitle=\bfseries,
    arc=2mm,
    boxrule=1.5pt,
    left=20pt,
    right=15pt,
]
\singlespacing
\vspace{-1cm}
\footnotesize
\begin{lstlisting}
Your task is to summarize the following question-answer pairs into a single string.
If the answer stated no information provided, only output "None" as your response; otherwise, output the summarized string.

{'Q: <Question1> A: <Answer1>', ...}
\end{lstlisting}

\end{tcolorbox}

\noindent
\underline{\textsc{Evaluator} agent}

The \textsc{Evaluator} agent assesses whether the final diagnosis is correct by comparing the predicted disease names with the ground truth, while accounting for variations in medical terminology and synonymous expressions. The following prompt is used:

\begin{tcolorbox}[
    colback=gray!10,
    colframe=gray!150,
    title={\textbf{\textsc{Evaluator} agent (to assess the correctness of diagnoses)}},
    fonttitle=\bfseries,
    arc=2mm,
    boxrule=1.5pt,
    left=20pt,
    right=15pt,
]
\singlespacing
\vspace{-1cm}
\footnotesize
\begin{lstlisting}
Evaluate the following diagnosis for correctness compared to the given ground truth.
You should be evaluating only the given diagnosis; you should not attempt to solve the task.
Respond: "true" if the diagnoses is correct, and "false" if the diagnosis are incorrect.

Ground truth: {}
Diagnosis to evaluate: {}
\end{lstlisting}
\end{tcolorbox}

\subsection*{Experimental setup}

Our experiments use a multi-agent simulation of clinical patient interviews. To reflect the diagnostic uncertainty from real-world clinical settings, we apply masking strategies to the patient cases, so that specific clinical information is hidden before the case is provided to the agents. We then evaluate all diagnostic systems on the same, masked (i.e., incomplete) patient cases. We use the following experiment designs: 
\begin{enumerate}
\item \textit{Single-feature masking:} To assess the diagnostic robustness to missing information, we mask one clinical feature category at a time in the structured case. Specifically, we mask one of the six categories: symptoms, social history, past medical history, physical examination, laboratory tests, or imaging results.
\item \textit{All-features masking:} To evaluate the information-seeking ability of a diagnostic system, we mask \textit{all} clinical feature categories in the patient case, except for the patient's demographics and the primary presenting symptom. In other words, all of the following categories are omitted at the same time: social history, past medical history, physical examination, laboratory tests, and imaging results. This setting is thus considerably more challenging than the single-feature masking.
\end{enumerate}
For comparison, we also include a baseline that operates on the full, unmasked patient case without any question asking, which serves as an upper bound on the performance of off-the-shelf LLMs. 

\textbf{Baselines.} We include the following baselines:
\begin{itemize}
    \item 
\textbf{Na{\"i}ve question generation:} Here, the \textsc{Doctor} agent generates follow-up questions without any explicit question selection via MedClarify. Instead, inspired by previous research \cite{spitzer_effect_2025}, the following prompt is used to generate differential diagnoses based on which the first question is used:

\begin{tcolorbox}[
    colback=gray!10,
    colframe=gray!150,
    title={\textbf{\textsc{Doctor} agent (to ask clinical question)}},
    fonttitle=\bfseries,
    arc=2mm,
    boxrule=1.5pt,
    left=20pt,
    right=15pt,
]
\singlespacing
\vspace{-1cm}
\footnotesize
\begin{lstlisting}[language=json, xleftmargin=0pt]
Your goal is to ask a question to rule out the <Differential diagnosis> based on <Task>.
The question should be bite-sized and not more than 1 sentence. Do not repeat the questions you have asked before in <Inquiries History>.
You are not allowed to ask directly if your diagnosis is correct or reveal your guess diagnosis explicitly.
Output your answer concisely in the following format:

Thought:
[Your thought process here]

Response:
[Your question here]
\end{lstlisting}

\end{tcolorbox}

\item \textbf{Question selection with na{\"i}ve entropy:} This refers to an ablation of MedClarify where DEIG is used without Div and Con components. In other words, this implements the MedClarify process from step 1 (assessment of candidate diagnoses), step 2 (question generation), and  step 3 (question selection), but using the standard expected information gain (EIG) formula:
\begin{equation}
    q^* = \arg\max_{q \in Q_t} \left[\mathrm{IG}(q)\right] .
\end{equation}
In contrast to the MedClarify system, the baseline obtains diagnostic probabilities directly from the LLM without performing Bayesian updates across iterations. The multi-agent architecture and all prompts are otherwise identical to MedClarify. Thereby, we can quantify the contribution of our diagnosis-tailored information gain to obtain case-specific follow-up questions.
\end{itemize}

% reporting

Each experiment is repeated across five runs with different random seeds to account for variability. Results are reported using mean and standard deviation across the five runs.

%%%%%%%%%%%%%%%%%%%%%%%%%%%%%%%%%%%%%%%%%%%%%%%%%%%%%%%%%%%%%%%%%%%%%%%%%%%%%%
\newpage

\vspace{0.4cm}
\section*{Data availability}

NEJM cases were downloaded from \url{https://www.nejm.org/image-challenge}, MediQ cases were downloaded from \url{https://github.com/stellalisy/mediQ}, and MedQA cases were downloaded from \url{https://github.com/jind11/MedQA}.

\vspace{0.4cm}
\section*{Code availability}

All code to replicate our analyses will be made publicly available upon acceptance.

%%%%%%%%%%%%%%%%%%%%%%%%%%%%%%%%%%%%%%%%%%%%%%%%%%%%%%%%%%%%%%%%%%%%%%%%%%%%%%

\newpage
\bibliography{literature}

%%%%%%%%%%%%%%%%%%%%%%%%%%%%%%%%%%%%%%%%%%%%%%%%%%%%%%%%%%%%%%%%%%%%%%%%%%%%%%

%%%%%%%%%%%%%%%%%%%%%%%%%%%%%%%%%%%%%%%%%%%%%%%%%%%%%%%%%%%%%%%%%%%%%%%%%%%%%%

\newpage
\section*{Acknowledgments}

SF acknowledges funding via the Swiss National Science Foundation (SNSF), Grant 186932.

\vspace{0.4cm}
\section*{Author contributions} 

H.M.W. implemented the code and performed the analysis. P.J. checked the code. H.M.W. and S.F. contributed to conceptualization and manuscript writing. P.H. and M.B. edited the draft. All authors approved the manuscript.

\vspace{0.4cm}
\section*{Competing interests}
The authors declare no competing interests.

\newpage

\appendix

\setcounter{figure}{1} % restart numbering
\renewcommand{\thefigure}{\arabic{figure}}
\renewcommand{\figurename}{Extended Figure}
%%%%%%%%%%%%%%%%%%%%%%%%%%%%%%%%%%%%%%%%%%%%%%%%%%%%%%%%
\section*{Supplementary Figures}

\begin{exlisting}[Structured patient case]\label{exlst:structured-patient-case}
    \singlespacing
    \vspace{-1cm}
    \footnotesize
    \begin{lstlisting}[language=json]
    {"Patient_Case": 
        {"Patient_Information":
            {"Demographics": "44-year-old man", 
             "History": ["The pain is felt in the retrosternal area and
                          radiates up to his left shoulder and arm", 
                         "The pain worsens on inspiration and is relieved
                          when he is leaning forward"], 
             "Symptoms": 
                {"Primary_Symptom": "sudden chest pain",
                 "Secondary_Symptoms": ["difficulty breathing"]}, 
             "Past_Medical_History": "", 
             "Social_History": "", 
             "Review_of_Systems": ""}, 
         "Physical_Examination": ["Heart rate is 61/min, respiratory
                                      rate is 16/min, temperature is 36.5C
                                      (97.7F), blood pressure is 115/78 mm
                                      Hg", 
                                      "Physical examination shows no
                                      abnormalities", 
                                      "Pericardial friction rub is heard on
                                      auscultation"], 
         "Test_Results": 
             {"Laboratory_Findings": ["elevated erythrocyte sedimentation rate
                              (ESR) and C-reactive protein (CRP)
                              levels"], 
              "Imaging_Results": [],
              "Other": ["An ECG is performed"]}}}
    \end{lstlisting}
\end{exlisting}

\newpage
\begin{figure}
\begin{center}
\includegraphics[width=.85\linewidth]{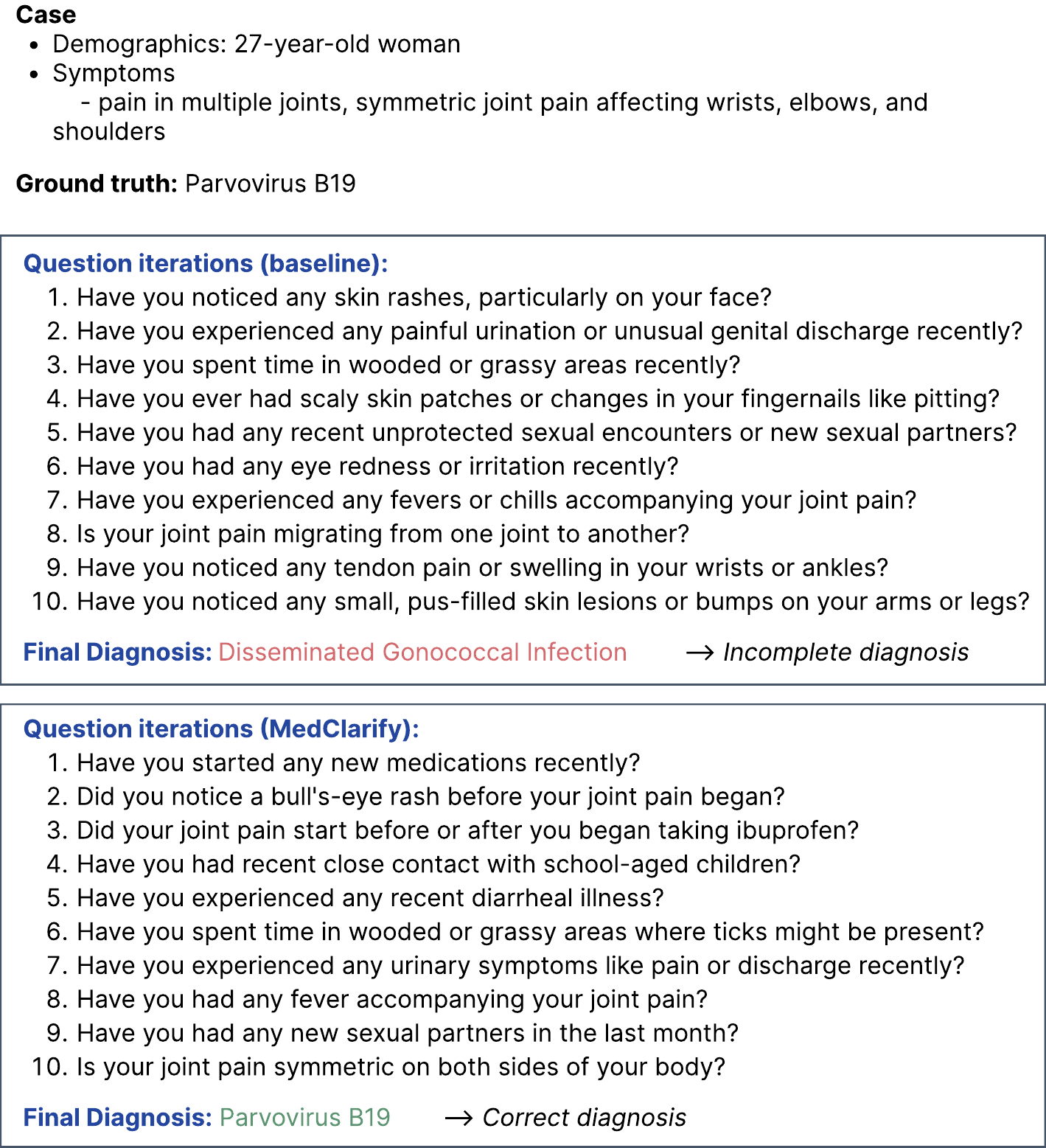}
\end{center}

\caption{\textbf{Example with proactive, case-specific follow-up questions for medical diagnosis.} The figure presents an example patient case with clinical information (top), as well an the diagnostic dialogue generated by the na{\"i}ve multi-turn baseline system (middle) and by MedClarify (bottom) for the same patient case.}
\label{exfig:ex-effective-ex1}
\end{figure}

\newpage
\begin{figure}
\begin{center}
\includegraphics[width=.85\linewidth]{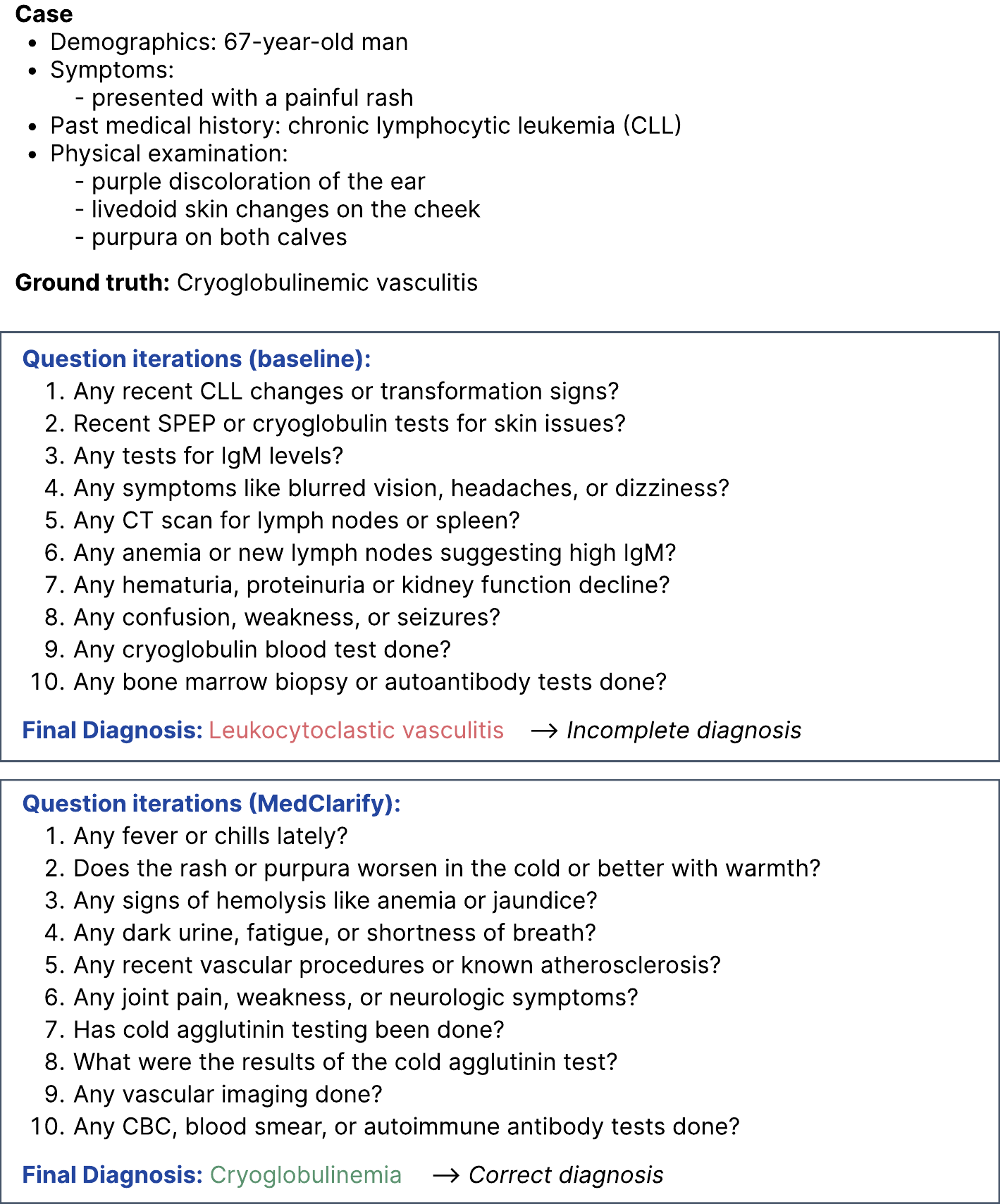}
\end{center}

\caption{\textbf{Example with proactive, case-specific follow-up questions for medical diagnosis.} The figure presents an example patient case with clinical information (top), as well an the diagnostic dialogue generated by the na{\"i}ve multi-turn baseline system (middle) and by MedClarify (bottom) for the same patient case.}
\label{exfig:ex-effective-ex2}
\end{figure}

\newpage
\begin{figure}
\begin{center}
\includegraphics[width=1\linewidth]{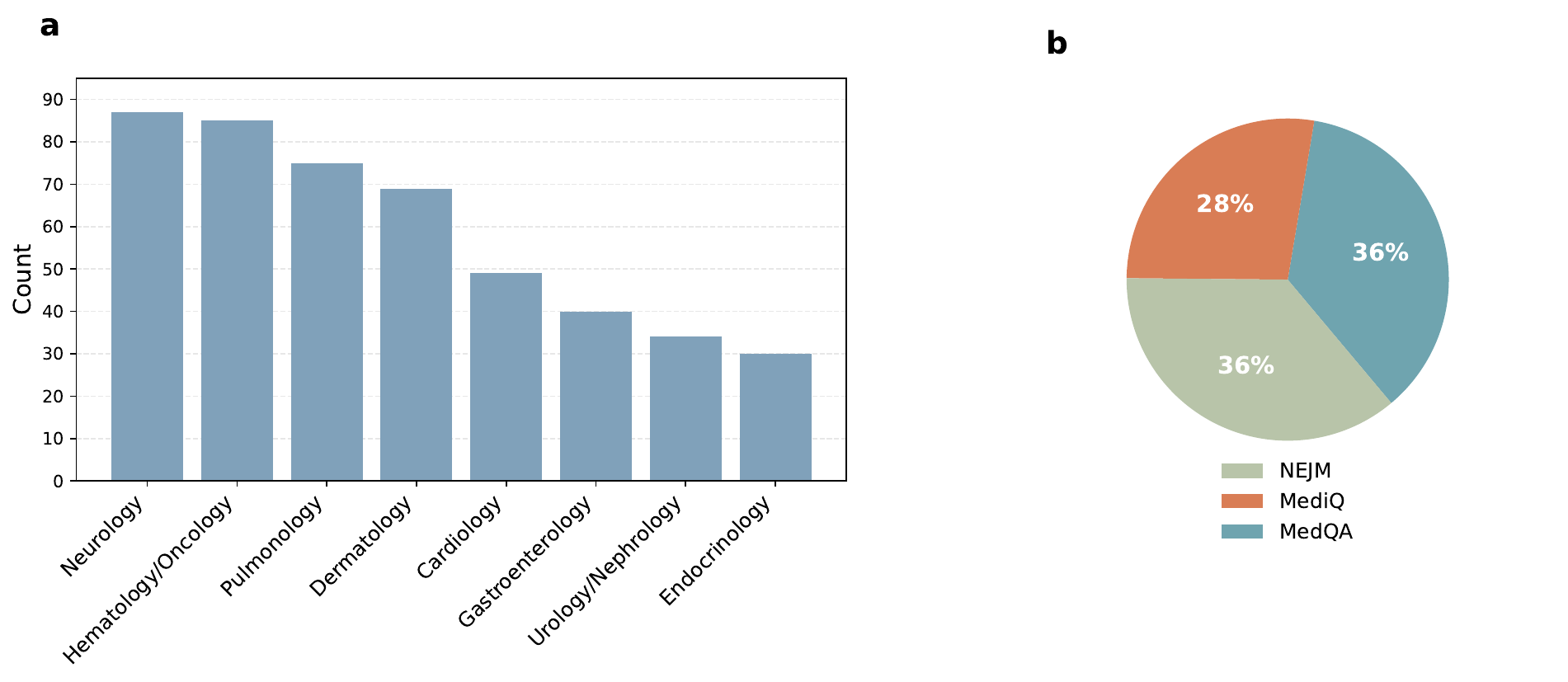}
\end{center}

\caption{\textbf{Dataset analysis.} \textbf{a},~Medical specialties across three datasets: NEJM, MediQ, and MedQA. \textbf{b},~Relative size of the datasets during the experiments. }
\label{fig:dist}
\end{figure}

\begin{figure}
\centering
\qquad\textbf{\textsf{b}}
\begin{center}
\vspace{-1cm}
\includegraphics[width=1\linewidth]{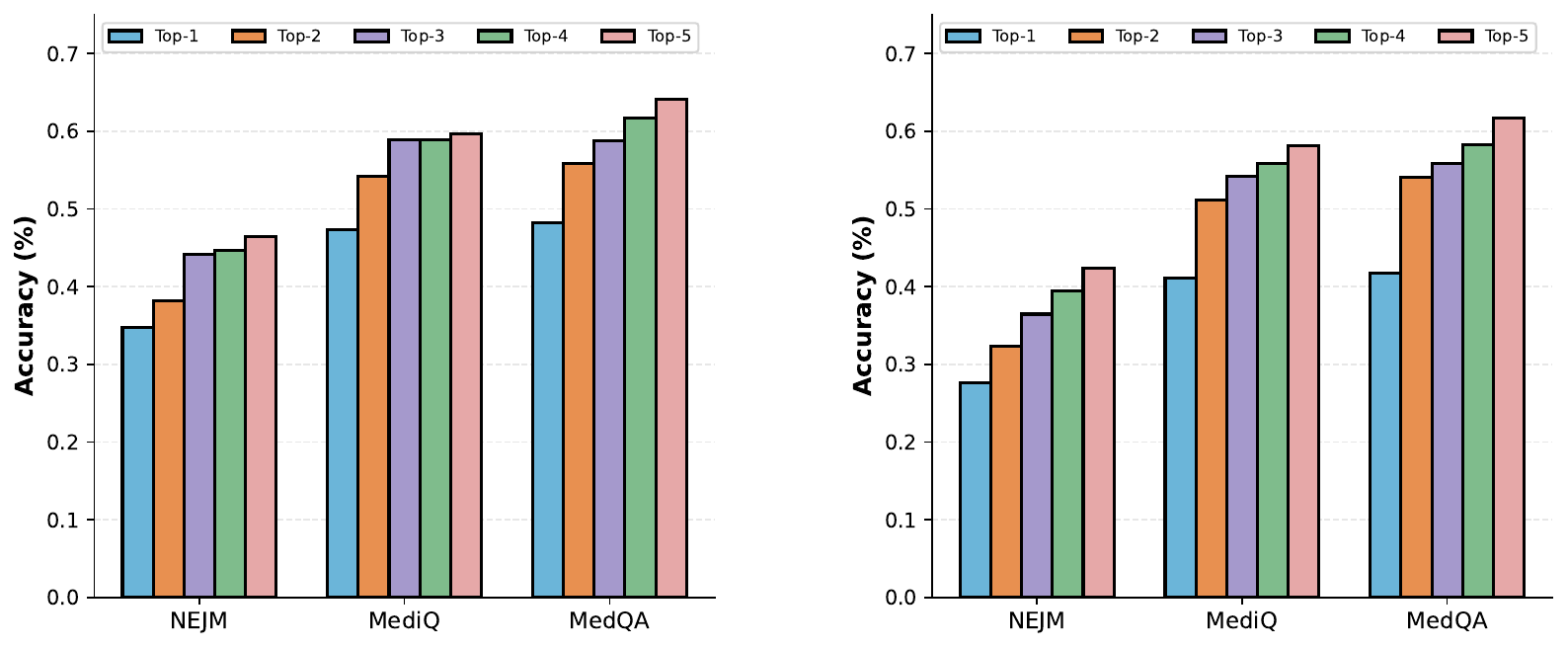}
\end{center}

\caption{\textbf{Overview of the diagnostic accuracy of MedClarify and baseline systems across datasets} The comparison of top-$k$ accuracy for MedClarify and baseline across NEJM, MediQ, and MedQA datasets in stacked bar charts.}
\label{exfig:accuracy}
\end{figure}

\newpage
\begin{figure}
\begin{center}
\includegraphics[width=.8\linewidth]{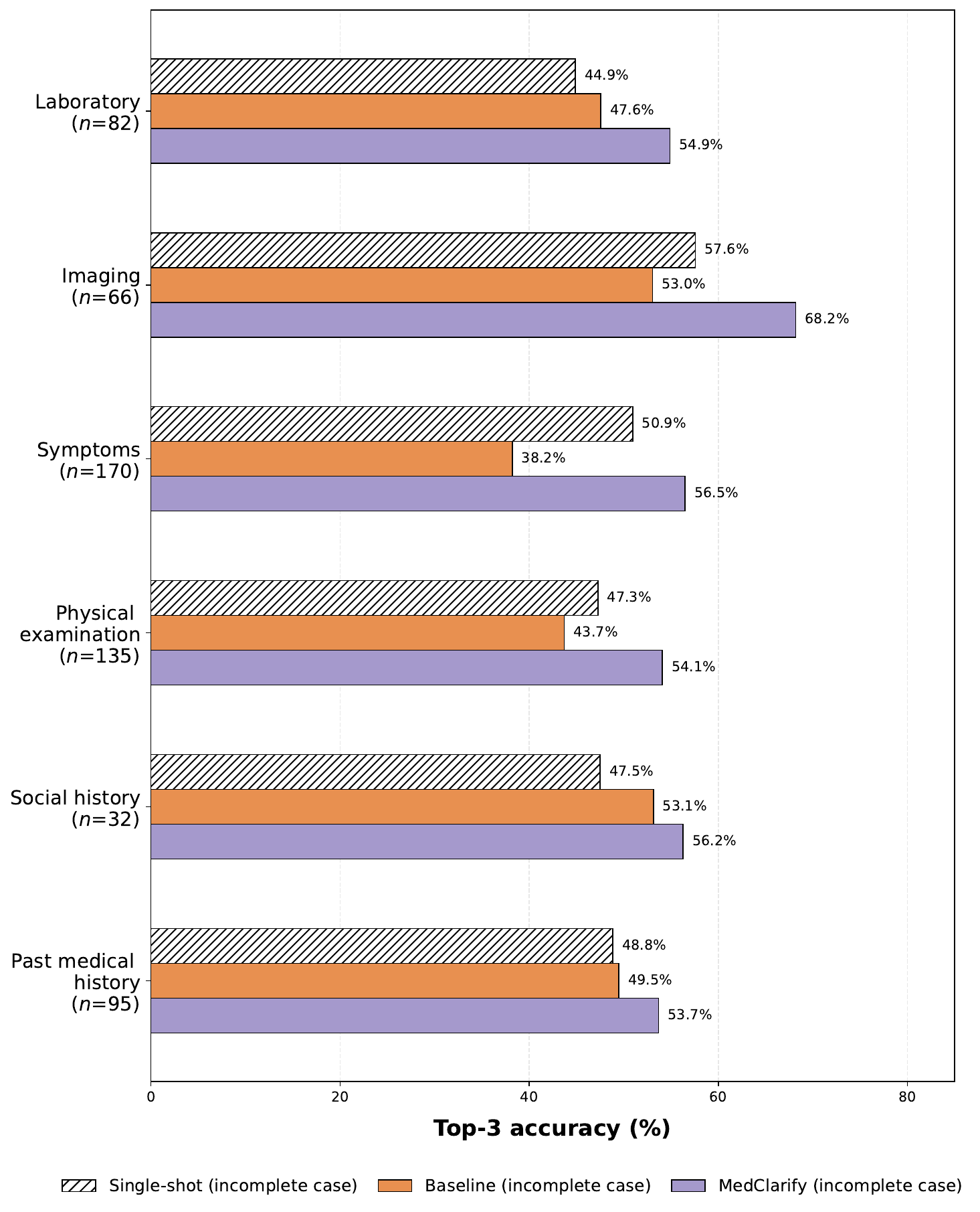}
\end{center}

\caption{\textbf{Results for top-3 accuracy showing how MedClarify improves diagnostic accuracy under incomplete patient cases.} Top-3 diagnostic accuracy on NEJM cases with one clinical feature category masked at a time (laboratory findings, imaging results, symptoms, physical examinations, social history, and past medical history), comparing a standalone single-shot LLM baseline (orange) with the MedClarify system (purple). The green bar indicates single-shot performance on complete cases and serves as an upper reference bound.}
\label{exfig:top-3-ablation}
\end{figure}

\newpage
\begin{figure}
\begin{center}
\includegraphics[width=0.86\linewidth]{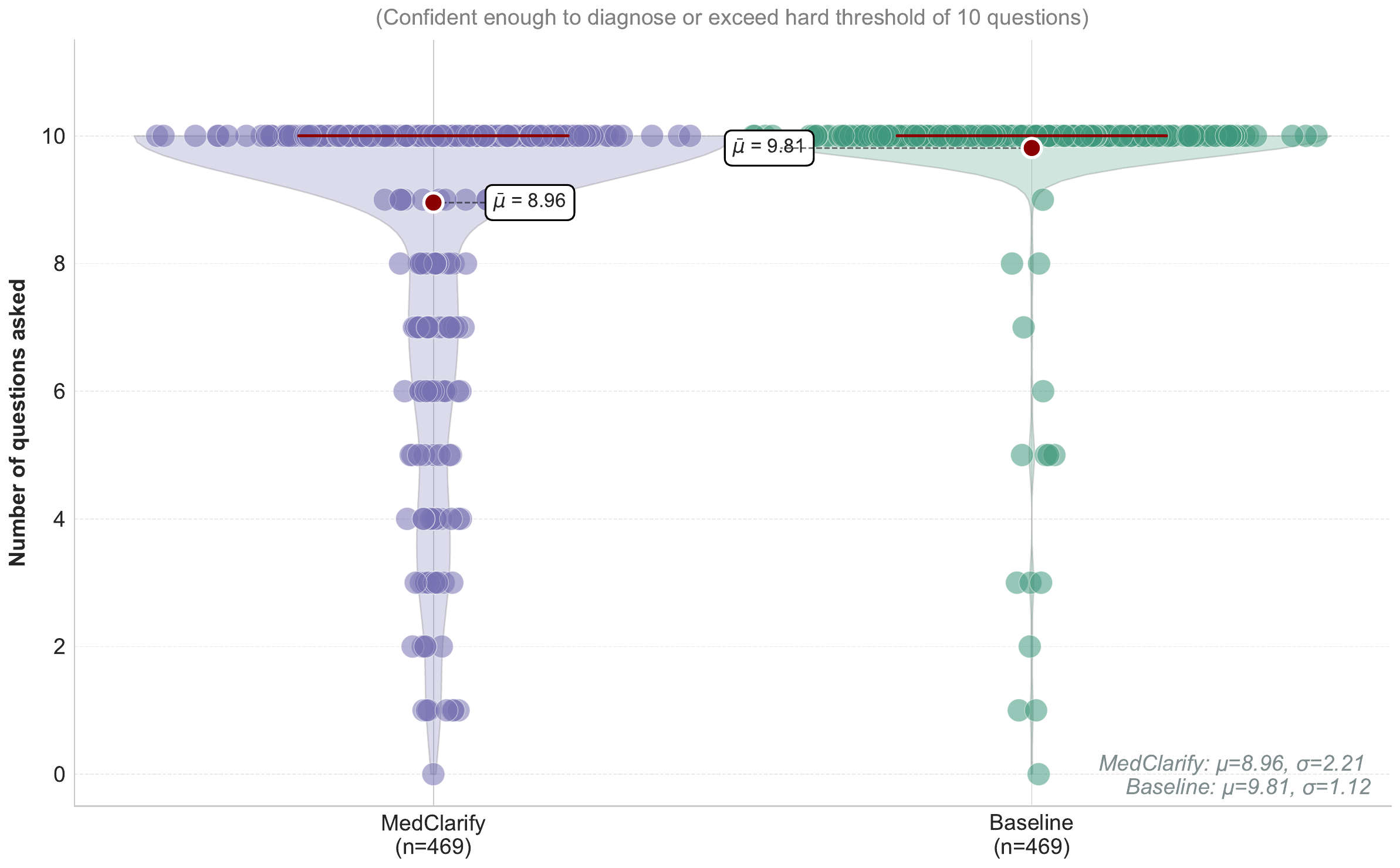}
\end{center}

\caption{\textbf{Efficiency of diagnostic information-seeking in MedClarify under joint masking of all feature categories.} Violin plot of the number of questions asked over iterations for MedClarify versus the na{\"i}ve multi-turn baseline to reach the correct diagnosis. Here, $\hat{\mu}$ denotes the mean number of questions per case. Fewer questions indicate more efficient acquisition of diagnostically relevant information through targeted follow-up.
}
\label{exfig:efficiency-a}
\end{figure}

\newpage
\begin{figure}
\begin{center}
\includegraphics[width=0.86\linewidth]{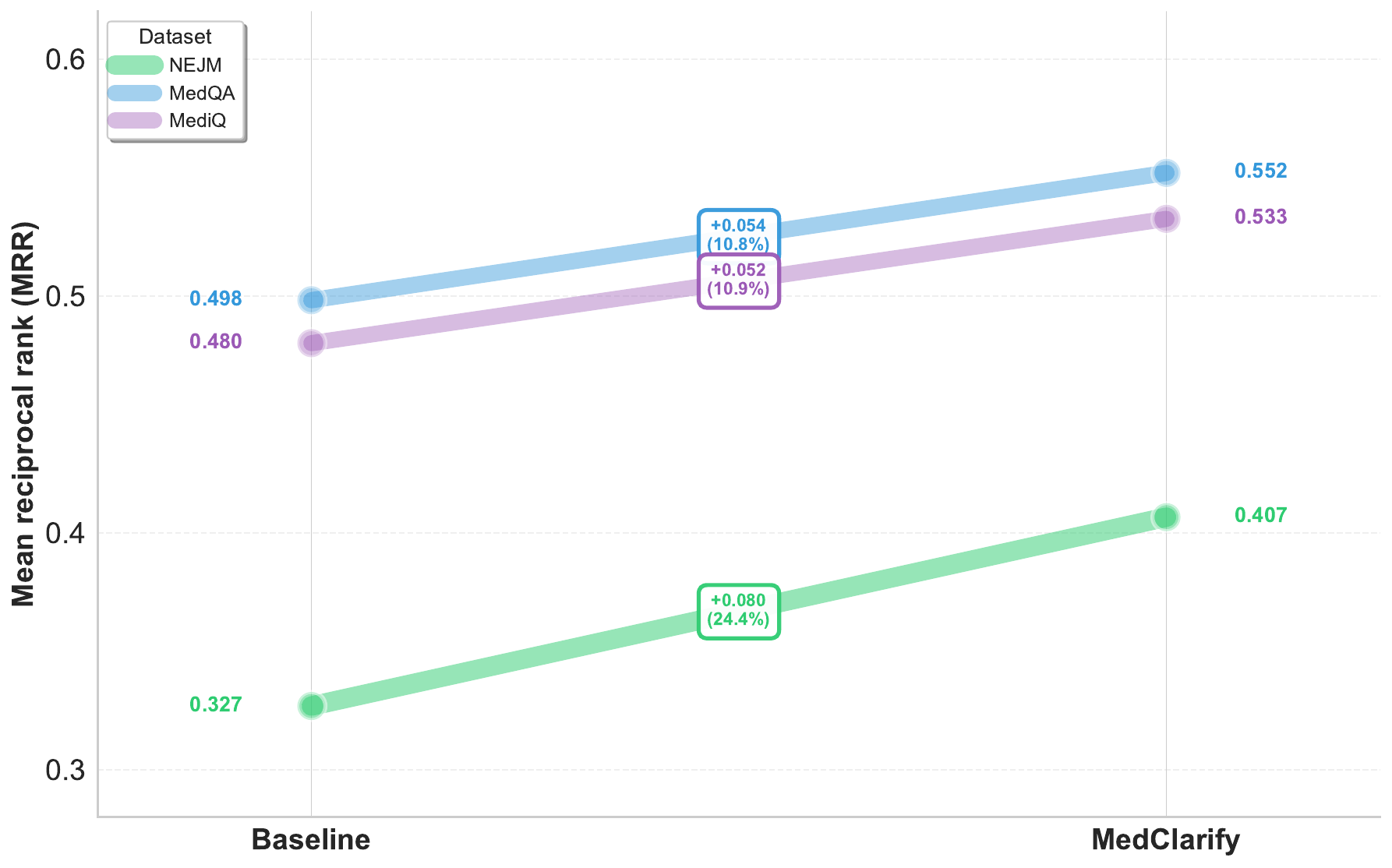}
\end{center}

\caption{\textbf{Efficiency of diagnostic information-seeking in MedClarify under joint masking of all feature categories.} Shown is the Mean Reciprocal Rank, which measures how early the correct diagnosis appears in the ranked differential diagnosis list. Higher MRR values indicate that MedClarify prioritizes the correct diagnosis earlier than the na{\"i}ve baseline system. 
}
\label{exfig:efficiency-b}
\end{figure}

\newpage
\begin{figure}
% \qquad\textbf{\textsf{a}}
\begin{center}
\includegraphics[width=1\linewidth]{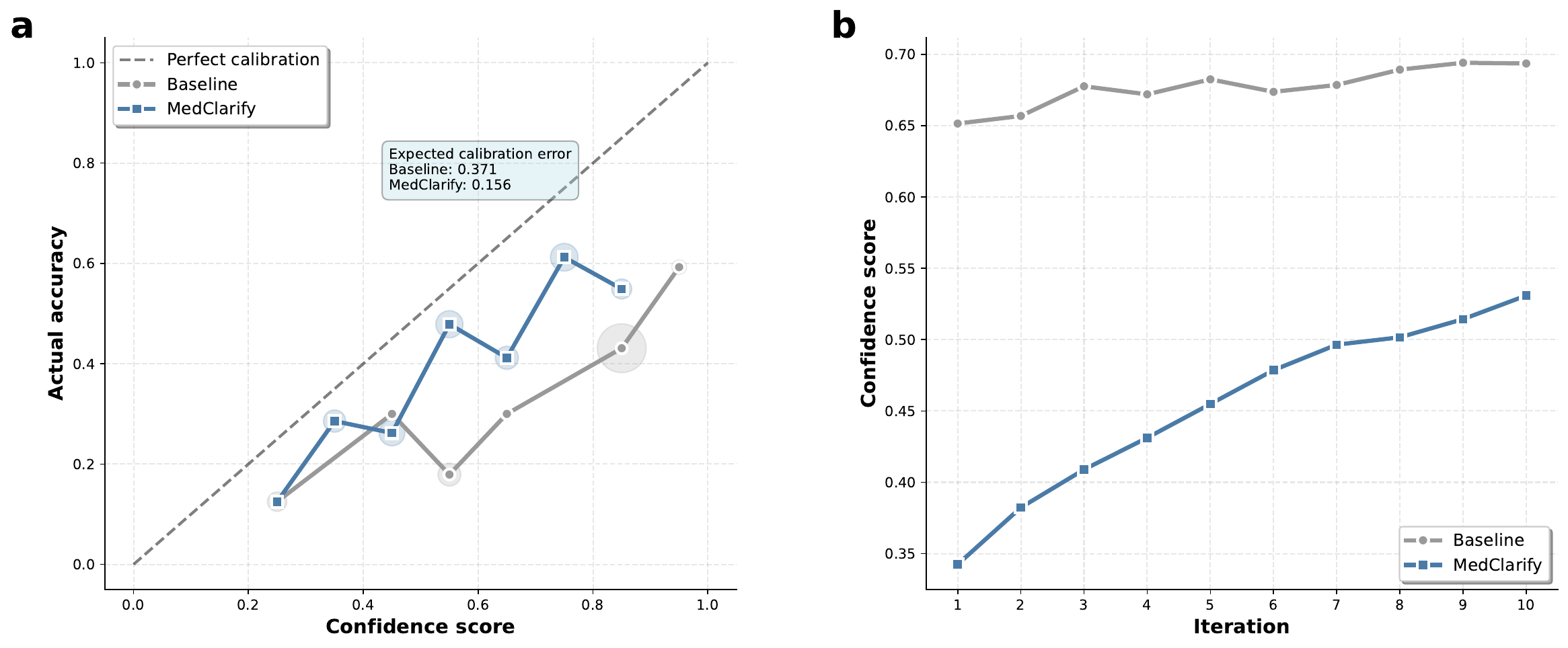}
\end{center}

\caption{\textbf{Confidence scores in MedClarify are clinically informative.} \textbf{a},~Evolution of confidence scores across diagnostic iterations comparing MedClarify with the na{\"i}ve multi-turn baseline. In the case of MedClarify, we see that the belief updates are reflected in increasing confidence scores as evidence accumulates. \textbf{b},~The calibration plot compares actual diagnostic accuracy with predicted confidence for each confidence bin. The dashed line represents perfect calibration. Together, these results show that MedClarify produces confidence estimates that are better aligned with diagnostic correctness, whereas the baseline exhibits higher but poorly calibrated confidence scores. }
\label{exfig:conf}
\end{figure}

\newpage
\begin{figure}
\centering
\llap{\textbf{\textsf{a}}\hspace{0.5em}}%
\includegraphics[width=.95\linewidth]{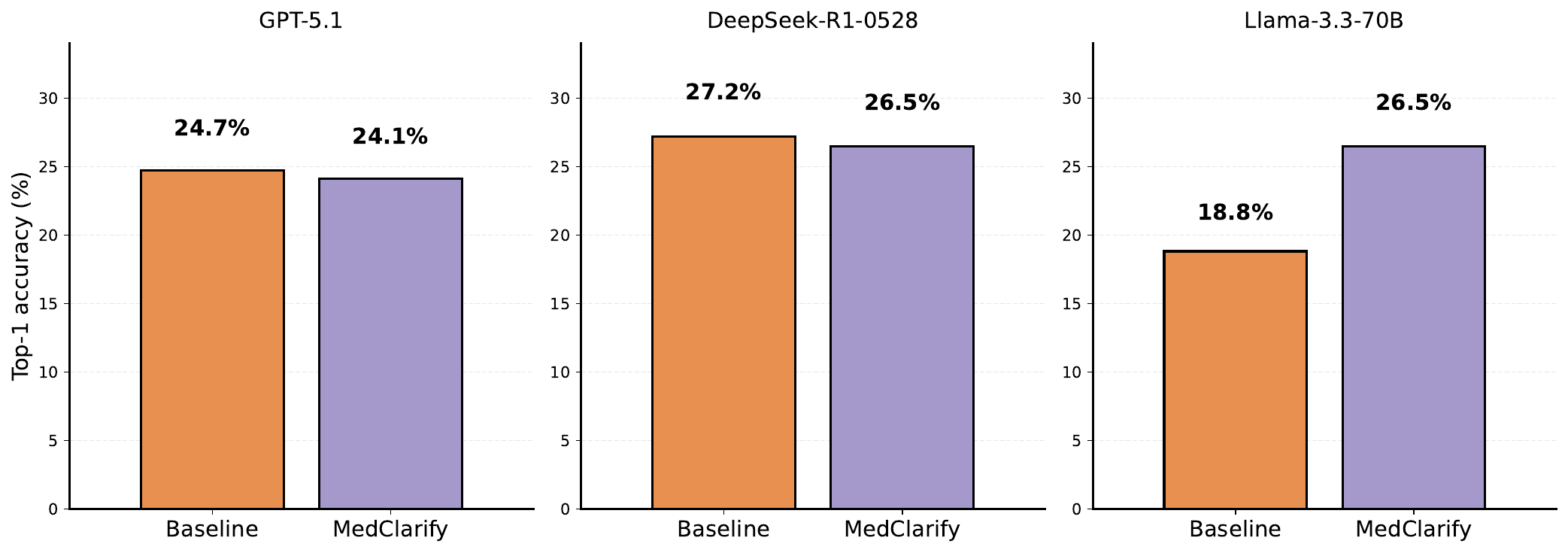}

\llap{\textbf{\textsf{b}}\hspace{0.5em}}%
\includegraphics[width=.95\linewidth]{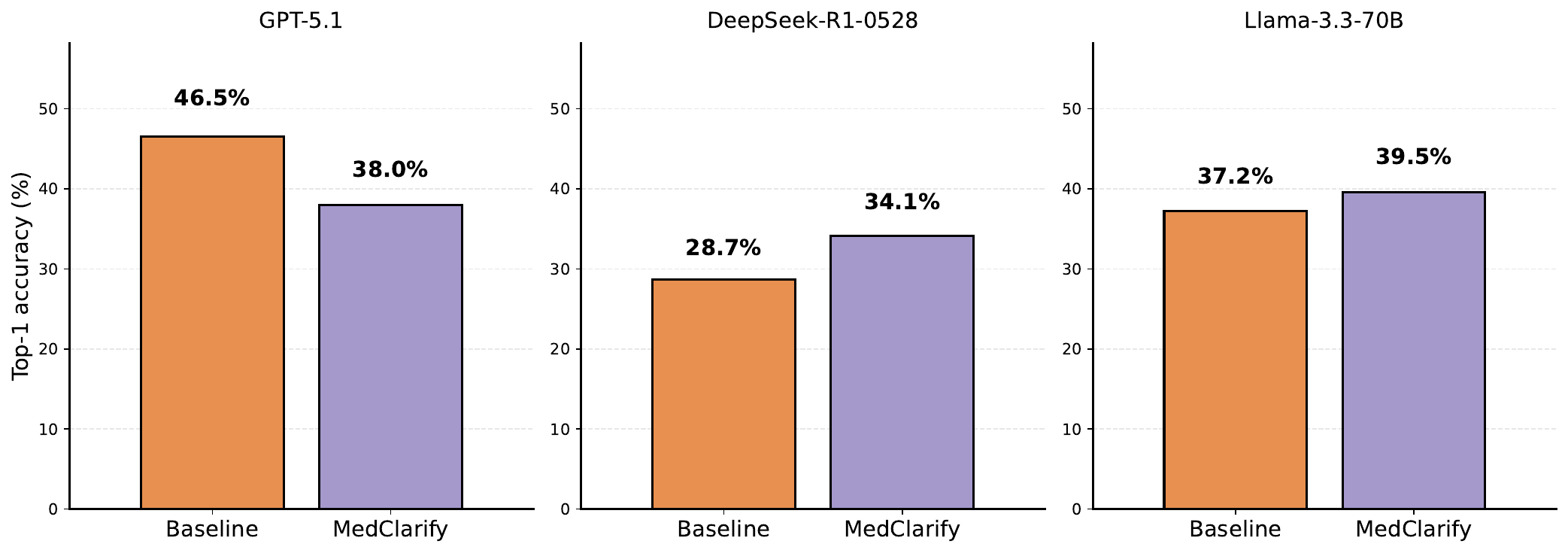}

\llap{\textbf{\textsf{c}}\hspace{0.5em}}%
\includegraphics[width=.95\linewidth]{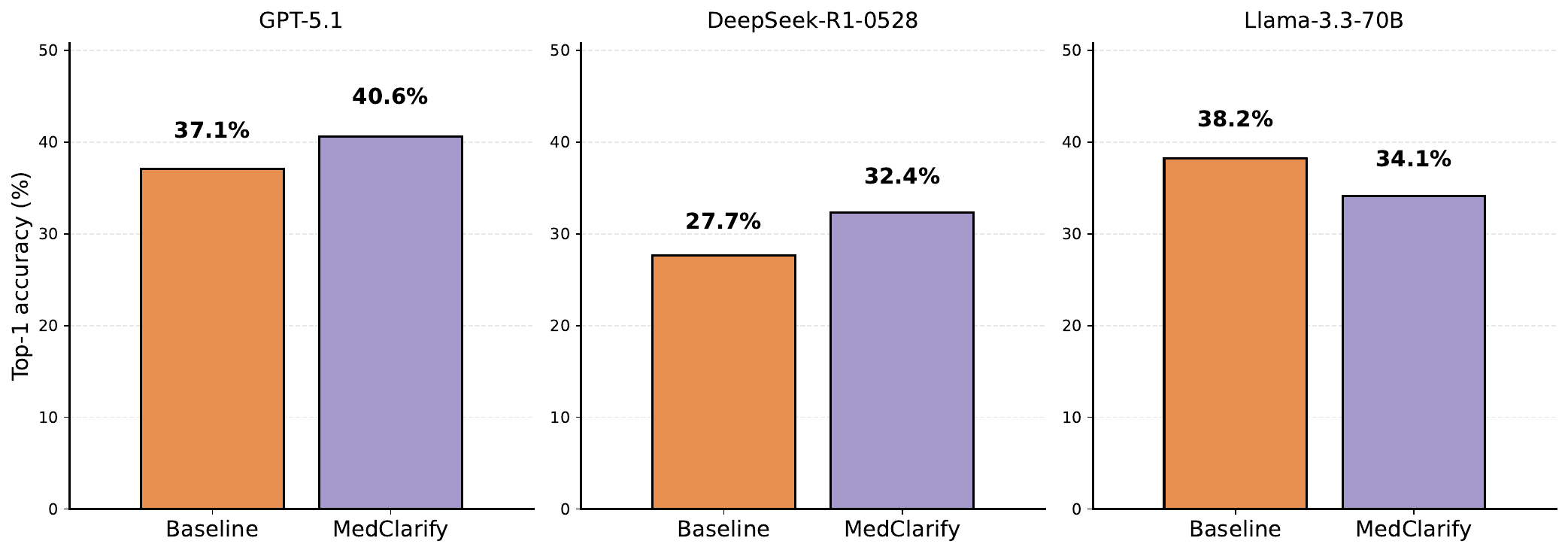}

\caption{\textbf{Overview of the diagnostic accuracy of MedClarify for up to five interaction questions across different LLM backbones.} Here, MedClarify is compared when instantiated with different LLM backbones, namely, GPT-5.1, Deepseek-R1-0528, and Llama-3.3-70B. Reported is the top-1 accuracy of MedClarify (purple) in comparison the na{\"i}ve multi-turn system across different datasets: \textbf{a},~NEJM; \textbf{b},~MediQ; and \textbf{c},~MedQA.}
\label{exfig:backbone-iter-five}
\end{figure}

\setcounter{table}{0} % restart numbering
\renewcommand{\thetable}{\arabic{table}}
\renewcommand{\tablename}{Extended Table}
%%%%%%%%%%%%%%%%%%%%%%%%%%%%
%%%%%%%%
\newpage
\section*{Supplementary Tables}
\begin{table}[h]
\centering
\caption{Performance comparison between baseline and MedClarify in terms of diagnostic accuracy in the all-features masking setting. A single-shot (incomplete case) indicates a direct diagnosis from the masked patient cases without follow-up questioning. A single-shot (complete case) indicates a direct diagnosis based on complete cases. The baseline is a na{\"i}ve multi-turn LLM system analogous to MedClarify but where questions are selected through a heuristic rather than an information-theoretic approach. The results are reported using 95\% confidence intervals, calculated from the sample mean and standard error as the mean plus or minus the margin derived from the Student's $t$-distribution. Abbreviation: s.e., standard error; CI, confidence interval.}
\label{tab:accuracy}
\begin{tabular}{lcccc}
\hline
\hline
Dataset & mean & s.e. & margin & 95\% CI \\
\hline
NEJM (single-shot complete)   & 0.368 & 0.005 & 0.007 & [0.362; 0.375] \\
MediQ (single-shot complete)  & 0.504 & 0.012 & 0.015 & [0.489; 0.519] \\
MedQA (single-shot complete)  & 0.499 & 0.013 & 0.016 & [0.483; 0.515] \\
\midrule
NEJM (single-shot incomplete) & 0.087 & 0.005 & 0.006 & [0.081; 0.093] \\
MediQ (single-shot incomplete) & 0.260 & 0.009 & 0.011 & [0.250; 0.271] \\
MedQA (single-shot incomplete) & 0.152 & 0.011 & 0.013 & [0.139; 0.165] \\
\midrule
NEJM (baseline)   & 0.280 & 0.003 & 0.008 & [0.272; 0.289] \\
MediQ (baseline)  & 0.412 & 0.017 & 0.018 & [0.395; 0.430] \\
MedQA (baseline)  & 0.443 & 0.023 & 0.024 & [0.419; 0.467] \\
\midrule
NEJM (MedClarify)   & 0.361 & 0.019 & 0.047 & [0.314; 0.408] \\
MediQ (MedClarify)  & 0.461 & 0.018 & 0.029 & [0.432; 0.491] \\
MedQA (MedClarify)  & 0.484 & 0.010 & 0.016 & [0.468; 0.500] \\
\hline
\multicolumn{4}{l}{$n$ = 170 (NEJM), 129 (MediQ), 170 (MedQA).} \\
\hline
\hline
\end{tabular}
\end{table}

\newpage

\begin{table}[h]
\centering
\caption{Dataset size across different feature categories. Values represent the number of cases containing each medical feature after preprocessing. We included only cases that focus on diagnostic reasoning with sufficient clinical details, excluding management, treatment, and prognosis content.}
\label{tab:medical-feature}
\begin{tabular}{l p{2.5cm} p{2.5cm} p{2.5cm}}
\hline
\hline
Feature category & \centering NEJM & \centering MediQ & \centering\arraybackslash MedQA \\
\hline
Symptoms & \centering 170 & \centering 129 & \centering\arraybackslash 170 \\
Past medical history & \centering 95 & \centering 88 & \centering\arraybackslash 170 \\
Social history & \centering 32 & \centering 71 & \centering\arraybackslash 170 \\
Physical examination & \centering 135 & \centering 118 & \centering\arraybackslash 170 \\
Laboratory findings & \centering 82 & \centering 58 & \centering\arraybackslash 170 \\
Imaging results & \centering 66 & \centering 34 & \centering\arraybackslash 170 \\
%\hline
%\multicolumn{4}{l}{Values represent the number of cases containing each medical feature.} \\
%\multicolumn{4}{l}{Total cases per dataset: NEJM ($n$ = 170), MediQ ($n$ = 129), MedQA ($n$ = 170)} \\
\hline
\hline
\end{tabular}
\end{table}

\end{document}